\documentclass{article} % For LaTeX2e
\usepackage{iclr2026_conference,times}
\usepackage{amsmath,amsfonts,bm}

\def\eqref#1{equation~\ref{#1}}
\def\1{\bm{1}}

\DeclareMathAlphabet{\mathsfit}{\encodingdefault}{\sfdefault}{m}{sl}
\SetMathAlphabet{\mathsfit}{bold}{\encodingdefault}{\sfdefault}{bx}{n}

\usepackage{hyperref}
\usepackage{enumerate}
\usepackage{url}
\usepackage{amsmath}
\usepackage{amssymb}
\usepackage{graphicx}
\usepackage{graphicx}
\usepackage{subcaption} % or subfig
\usepackage{tcolorbox}
\usepackage{hyperref}
\usepackage{xcolor}
\usepackage{cancel}
\usepackage{booktabs}
\usepackage{multirow}

\newtheorem{theorem}{Theorem}

\title{Unlocking the Power of Multi-Agent LLM for Reasoning: From Lazy Agents to Deliberation}

\author{
Zhiwei Zhang$^{1}$, \ Xiaomin Li$^{2}$, \ Yudi Lin$^{3}$, \ Hui Liu, \ Ramraj Chandradevan, \ Linlin Wu$^{4}$,\\
\textbf{Minhua Lin}$^{1}$, \ \textbf{Fali Wang}$^{1}$, \ \textbf{Xianfeng Tang}, \ \textbf{Qi He}$^{5}$, \ \textbf{Suhang Wang}$^{1}$\\[4pt]
$^{1}$The Pennsylvania State University \
$^{2}$Harvard University \ 
$^{3}$Michigan State University \ \\
$^{4}$University of Utah \
$^{5}$Microsoft \\
\texttt{\{zbz5349, szw494\}@psu.edu}
}

\iclrfinalcopy % Uncomment for camera-ready version, but NOT for submission.
\begin{document}

\maketitle

\begin{abstract}
Large Language Models (LLMs) trained with reinforcement learning and verifiable rewards have achieved strong results on complex reasoning tasks. Recent work extends this paradigm to a multi-agent setting, where a meta-thinking agent proposes plans and monitors progress while a reasoning agent executes subtasks through sequential conversational turns. Despite promising performance, we identify a critical limitation: lazy agent behavior, in which one agent dominates while the other contributes little, undermining collaboration and collapsing the setup to an ineffective single agent.  In this paper, we first provide a theoretical analysis showing why lazy behavior naturally arises in multi-agent reasoning. We then introduce a stable and efficient method for measuring causal influence, helping mitigate this issue. Finally, as collaboration intensifies, the reasoning agent risks getting lost in multi-turn interactions and trapped by previous noisy responses. To counter this, we propose a verifiable reward mechanism that encourages deliberation by allowing the reasoning agent to discard noisy outputs, consolidate instructions, and restart its reasoning process when necessary. Extensive experiments demonstrate that our framework alleviates lazy agent behavior and unlocks the full potential of multi-agent framework for complex reasoning tasks.

\end{abstract}

\section{Introduction}

% \zhiwei{[xiaomin writes a paragraph of background on LLM reasoning]}
Recent advances in prompting and training have markedly improved the multi-step reasoning abilities of large language models (LLMs) \citep{wei2022chain,kojima2022large,wang2022self,zhang2022automatic,ton2024understanding,yeo2025demystifying,zhu2025chain,chowdhury2025zero,mukherjee2025premise,balcan2025learning}. Techniques such as chain-of-thought prompting \citep{wei2022chain,kojima2022large} and structured methods like Tree-of-Thoughts and Graph-of-Thoughts \citep{yao2023tree,besta2024graph} expand the space for deliberation. Building on this, Large Reasoning Models (LRMs) trained with supervised and reinforcement learning using verifiable rewards achieve strong performance on math, code, and planning tasks \citep{jaech2024openai,guo2025deepseek,comanici2025gemini,plaat2024reasoning,huang2022towards,zhang2023cumulative,li2025webthinker,chen2025towards}. More recently, multi-agent frameworks enable LLMs with specialized roles to collaborate via planning, delegation, and debate, echoing human team dynamics \citep{li2023camel,wu2024autogen,chen2023autoagents,du2023improving,yuanreinforce}. Likewise, single-agent multi-turn interaction settings have gained attention as another path to enhance reasoning \citep{wan2025rema,shi2025mobilegui,wei2025webagent,zhou2025sweet,li2025beyond,lu2025deepdive,wang2025ragen,zhang2025survey,zeng2025reinforcing,jin2025search}.

% Recent advances in prompting and training large language models (LLMs) have significantly improved their multi-step reasoning abilities \citep{wei2022chain,kojima2022large,wang2022self,zhang2022automatic,ton2024understanding,yeo2025demystifying,zhu2025chain,chowdhury2025zero,mukherjee2025premise,balcan2025learning}. Techniques like chain-of-thought prompting \citep{wei2022chain, kojima2022large} and structured methods such as Tree-of-Thoughts and Graph-of-Thoughts \citep{yao2023tree, besta2024graph} have expanded the space for deliberation. Building on this, Large Reasoning Models (LRMs) trained with supervised and reinforcement learning with verifiable rewards achieve strong performance on math, code, and planning tasks \citep{jaech2024openai, guo2025deepseek, comanici2025gemini, plaat2024reasoning, huang2022towards, zhang2023cumulative, li2025webthinker, chen2025towards}. More recently, multi-agent frameworks have emerged, where LLMs with specialized roles collaborate via planning, delegation, and debate to tackle complex reasoning problems, echoing human team dynamics \citep{li2023camel, wu2024autogen, chen2023autoagents, du2023improving, yuanreinforce}. Similarly, single-agent settings with multi-turn interactions have also attracted considerable attention \citep{wan2025rema, shi2025mobilegui, wei2025webagent, zhou2025sweet, li2025beyond, lu2025deepdive, wang2025ragen, zhang2025survey, zeng2025reinforcing, jin2025search}.

To support multi-agent and multi-turn reinforcement learning, multi-turn Group Relative Preference Optimization (GRPO) \citep{wan2025rema,shi2025mobilegui,wei2025webagent} and its variants~\citep{guo2025segment,zhang2025stephint,ning2025not,xue2025simpletir} compute advantages and importance ratios at the turn level, enabling finer-grained optimization and more precise credit assignment. Building on this foundation, ReMA \citep{wan2025rema} introduces a multi-agent LLM reasoning framework with two specialized roles: a \emph{meta-thinking agent}, which decomposes tasks, sets intermediate goals, and adapts based on feedback, and a \emph{reasoning agent}, which performs step-by-step computations and proofs before returning intermediate results. The agents alternate sequentially, but since only a final outcome reward is available, ReMA computes a group advantage following GRPO \citep{shao2024deepseekmath} and uniformly assigns this trajectory-level signal to every turn in the rollout.

% Despite its effectiveness, we empirically find that ReMA suffers from a critical issue of lazy agents, where one of the agents contributes only trivially to the system. While this phenomenon is well documented in traditional multi-agent RL under sparse rewards \citep{sunehag2018value,foerster2018counterfactual,castellini2022difference,jaques2019social,wang2020roma,liu2023lazy}, prior work has focused on settings where agents act simultaneously. \emph{By contrast, our findings are surprising because agents in our setting act sequentially:} an early agent’s trivial action not only fails to contribute but also alters the state for subsequent agents. Since later decisions build on this evolving state, such laziness can misguide the reasoning trajectory and amplify its negative influence. Intuitively, such interdependence should discourage laziness as training improves overall performance, yet we observe that ReMA-trained reasoning agents still adopt shortcut behaviors.  
% As shown in Section~\ref{sec:preliminary}, our case study reveals that reasoning agents often contribute only superficially—summarizing or copying the meta-thinking agent’s responses rather than engaging in genuine questioning or reflection. Consequently, the meta-thinking agent bears nearly the entire reasoning burden. Causal-effect experiments further confirm that although both agents initially contribute when initialized from the base model, the reasoning agent’s influence declines sharply with training, leaving the meta-thinking agent dominant.

Despite its effectiveness, we empirically find that ReMA suffers from a critical issue of lazy agents: one of the agents contributes only trivially to the multi-agent system.
% As shown in Fig.~\ref{fig:framework} (a), the lazy reasoning agent leaves the intermediate steps blank and only provides a final summary, leading to incorrect results. \zhiwei{check whether can add more later}
Although this phenomenon has been widely acknowledged in traditional multi-agent reinforcement learning under sparse-reward settings \citep{sunehag2018value, foerster2018counterfactual, castellini2022difference, jaques2019social, wang2020roma, liu2023lazy}, prior work has primarily focused on scenarios where multiple agents act simultaneously. 
\emph{In contrast, our findings are surprising because agents in our setting act sequentially.} An early agent’s trivial action not only fails to contribute but also shapes the state for subsequent agents. As later decisions depend on this evolving state, a lazy action can misguide the reasoning trajectory and compound its negative influence. Intuitively, such interdependence across turns should discourage laziness, especially as overall performance improves during training. However, contrary to this expectation, we find that ReMA-trained reasoning agents still adopt shortcut behaviors.
As shown in Section~\ref{sec:preliminary}, our case study reveals that reasoning agents often contribute only trivially, typically by summarizing or copying the meta-thinking agent’s responses without genuine questioning or reflection. As a result, the meta-thinking agent ends up carrying almost the entire reasoning process. Our causal-effect experiments further show that while both agents initially contribute substantially when initialized from the base model, the reasoning agent’s influence diminishes markedly as training progresses, leaving the meta-thinking agent dominant.
The critical issue of lazy agents in multi-agent systems risks collapsing the entire system into a single agent, thereby limiting the potential benefits of collaboration in improving performance.
In this paper, we propose Multi-Agent Meta-Reasoning Done Right \textbf{(Dr. MAMR)}. We begin with a theoretical analysis of multi-turn GRPO to investigate the root cause of lazy agent behavior and identify a key bias in its loss formulation: the normalization term, intended to prevent sequence-level bias toward longer rollouts, inadvertently drives the model to prefer continuations that minimize the number of turns given the same prefix. As a result, agents are implicitly incentivized to complete reasoning with fewer interactions, often bypassing collaborative reflection or correction, and over time, this dynamic gives rise to lazy agents that contribute little to the reasoning process.
Our theoretical insight not only explains the emergence of lazy agents but also sheds light on future work in designing objectives for multi-turn reinforcement learning.

While correcting the loss formulation partially mitigates the problem, it does not eliminate it. To further address this issue, we propose measuring the causal influence \citep{bogdan2025thought} of each reasoning step on subsequent process. 
A challenge arises in online training: the policy generates only a single continuation per step, so the estimated influence reflects just one trajectory. In contrast, considering multiple continuations would show how the step contributes across diverse trajectories, providing a more reliable estimate of its overall contribution and mitigating potential bias introduced by phrasing \citep{pavlick2016simple, mccoy2019right, merrick2020explanation, li2024shapley}, but such resampling is computationally prohibitive in online RL.
Inspired by \citet{feng2025group, li2021shapley}, we introduce a Shapley-inspired causal influence method. Instead of evaluating each step in isolation, we group semantically similar steps across rollouts and average their influence scores. This avoids additional sampling and produces robust estimates during training.  

As lazy behavior diminishes and agents engage more productively, interaction frequency increases. However, as \citet{laban2025llms} show, LLMs in multi-turn settings often overcommit to incomplete early context, making premature assumptions. A similar risk arises here: the meta-thinking agent acts like a user providing incremental instructions, while the reasoning agent may become misled by its own earlier outputs, as confirmed in Sec. ~\ref{sec:deliberation}. 
To overcome this, we propose training the reasoning agent to adaptively discard its prior outputs, re-aggregate instructions, and restart reasoning when needed. To accurately credit such restart behavior, we design a novel verifiable reward mechanism. Building on this, we assign step-level credit by aggregating outcome reward, causal influence, and restart signals. Extensive experiments demonstrate that our method effectively mitigates lazy-agent behavior and unlocks the potential of multi-agent frameworks for complex reasoning.

Overall, our contributions are as follows:
\textbf{(1)} We identify a critical issue of lazy agents in multi-agent reasoning frameworks and provide a theoretical analysis of multi-turn GRPO to explain its underlying cause.
\textbf{(2)} We propose a Shapley-inspired method for measuring causal influence at the step level, further mitigating the lazy agent problem.
\textbf{(3)} As agents engage in more frequent collaboration, we design a novel verifiable reward mechanism for restart behavior, enabling the reasoning agent to recover from noisy intermediate steps and avoid getting lost in prolonged interactions, thereby pushing the performance boundary of multi-agent LLMs on complex reasoning tasks.

\section{Related work}
\noindent\textbf{Multi-Agent RL.} Multi-agent RL (MARL) studies how agents coordinate to maximize collective rewards, with credit assignment as a central challenge. Classical approaches include value decomposition \citep{sunehag2018value}, counterfactual baselines \citep{foerster2018counterfactual}, regression-based rewards \citep{castellini2022difference}, role-based coordination \citep{wang2020roma}, and model-based influence estimation \citep{liu2023lazy}. With LLM agents, MARL has been adapted for multi-turn reasoning and dialogue, e.g., turn-level credit assignment \citep{zeng2025reinforcing}, critic-driven step-wise rewards \citep{zhou2025sweet}, communication-efficient training \citep{liao2025marft}, addressing coarse reward traps \citep{wang2025ragen}, and framing LLM collaboration via MAGRPO \citep{liu2025llm}. A persistent issue is lazy agents, motivating causal influence estimation \citep{bogdan2025thought, nguyen2025causalplan, liu2024large} to enable finer-grained credit assignment.

\noindent\textbf{LLM reasoning.} Large Language Models (LLMs) excel across diverse NLP tasks \citep{brown2020language, chowdhery2023palm, du2022glam, dubey2024llama, wenzek2019ccnet}. Chain-of-thought prompting improves reasoning by eliciting intermediate steps \citep{wei2022chain, kojima2022large, nye2021show}, while extensions like Tree-of-Thoughts and Graph-of-Thoughts enable structured, non-linear reasoning \citep{yao2023tree, besta2024graph}. These advances motivate Large Reasoning Models (LRMs) \citep{guo2025deepseek, achiam2023gpt, grattafiori2024llama, xu2023re, zhou2022least, wu2024comparative, qi2024mutual, chae2024language}, which combine supervised fine-tuning and reinforcement learning to achieve state-of-the-art results on math, coding, and planning \citep{jaech2024openai, guo2025deepseek, comanici2025gemini, yang2024qwen2, yang2025qwen3, lightman2023let, wang2023math}. Beyond single-model reasoners, multi-agent frameworks leverage role assignment, orchestration, and debate to coordinate specialized LLM agents for complex tasks \citep{li2023camel, wu2024autogen, chen2023autoagents, du2023improving, yuanreinforce}.

A detailed review of related work on MARL, hierarchical RL and LLM reasoning is in Appendix \ref{app:relatedwork}.

\section{Background}
\label{sec:background}

ReMA~\citep{wan2025rema} models reasoning as a \emph{multi-turn meta-thinking process} defined as:  
\begin{equation}
    \mathbf{x} \xrightarrow[\pi_h]{\text{meta-thinking}} \mathbf{m}_1
    \xrightarrow[\pi_l]{\text{reasoning}} \mathbf{y}_1 
    \xrightarrow[\pi_h]{\text{meta-thinking}} \mathbf{m}_2
    \xrightarrow[\pi_l]{\text{reasoning}} \mathbf{y}_2 
    \dots 
    \xrightarrow[\pi_l]{\text{reasoning}} \mathbf{y}_T,
    \label{eq:mamrp2}
\end{equation}
where $T$ is the number of turns. The high-level policy $\pi_h$ (\emph{meta-thinking agent}) generates meta-level thoughts $\mathbf{m}_t$ from the input $\mathbf{x}$ and history $\{\mathbf{m}, \mathbf{y}\}_{<t}$, while the low-level policy $\pi_l$ (\emph{reasoning agent}) produces token-level outputs $\mathbf{y}_t$ under the guidance of $\mathbf{m}_t$. To improve training efficiency, both agents share the same model weights $\theta$ but are distinguished by role-specific system prompts $S_h$ and $S_l$: $
\pi_h = \pi_\theta(\cdot \mid S_h, \cdot), \pi_l = \pi_\theta(\cdot \mid S_l, \cdot).
$

\textbf{Multi-turn GRPO} \citep{wan2025rema, wei2025webagent, shi2025mobilegui} extends GRPO \citep{guo2025deepseek} to support end-to-end multi-turn tasks such as mathematical reasoning \citep{wan2025rema} and web-based agent decision-making \citep{wei2025webagent}. A key innovation is the introduction of a turn-level importance ratio, enabling fine-grained credit assignment across dialogue turns.

% \citep{zhou2025sweet} also introduces turn-wise advantage function.

Specifically, given the dataset $\mathcal{D}$ and $G$ trajectories for each question, the objective is defined as:
{\small
\begin{equation}
\begin{aligned}
\label{eq:rema}
&\mathcal{J}(\theta) = \mathbb{E}_{(\mathbf{x},\mathbf{y}^*)\sim\mathcal{D},\, \{(\mathbf{m}_i, \mathbf{y}_i)\}_{i=1}^G \sim \pi_{\theta_{\mathrm{old}}}(\cdot\mid\mathbf{x})} 
\\
&\left[
\frac{1}{G} \sum_{i=1}^G \frac{1}{T_i} \sum_{t=1}^{T_i} \frac{1}{|\mathbf{y}_{i,t}|} \sum_{j=1}^{|\mathbf{y}_{i,t}|}
\left(
\min\left(r_{i,t}(\theta) \hat{A}_{i,t,j},\, \mathrm{clip}(r_{i,t}(\theta), 1-\epsilon, 1+\epsilon)\hat{A}_{i,t,j}\right)
- \beta\, D_{\mathrm{KL}}(\pi_\theta \| \pi_{\mathrm{ref}})
\right)
\right],
\end{aligned}
\end{equation}
}%
where $\mathbf{y}_{i,t,j}$ denotes the $j$-th token at turn $t$ in trajectory $i$, $\hat{A}_{i,t,j}$ is the token-level advantage and $\frac{1}{T_i}$ is a normalization to avoid bias toward rollouts with more turns. The turn-level importance ratio $r_{i,t}(\theta)$ is computed as:
\begin{align}
    r_{i,t}(\theta)
    &= \frac{1}{|\mathbf{y}_{i,t}|} \sum_{j=1}^{|\mathbf{y}_{i,t}|}
    \frac{
    \pi_\theta
    \bigl(
        \mathbf{y}_{i,t,j}\mid \mathbf{x}, \{\mathbf{m}_{i,\cdot}, \mathbf{y}_{i,\cdot}\}_{<t},  \mathbf{m}_{i, t}, \mathbf{y}_{i,t,<j}
    \bigr)
    }{
    \pi_{\theta_{\mathrm{old}}}
        \bigl(
            \mathbf{y}_{i,t,j}\mid \mathbf{x}, \{\mathbf{m}_{i,\cdot}, \mathbf{y}_{i,\cdot}\}_{<t},  \mathbf{m}_{i, t}, \mathbf{y}_{i,t,<j}
        \bigr)
    },
    \label{eq:turn_grpo}
\end{align}
which aggregates token-level likelihood ratios within each reasoning turn. Similar variants of Eq.~\ref{eq:rema} were also proposed in \citep{guo2025segment, zhang2025stephint, ning2025not, xue2025simpletir}.

\section{The lazy agent issue in multi-agent LLM reasoning}
\label{sec:preliminary}
\begin{figure}[t]
    \centering    \includegraphics[width=1\linewidth]{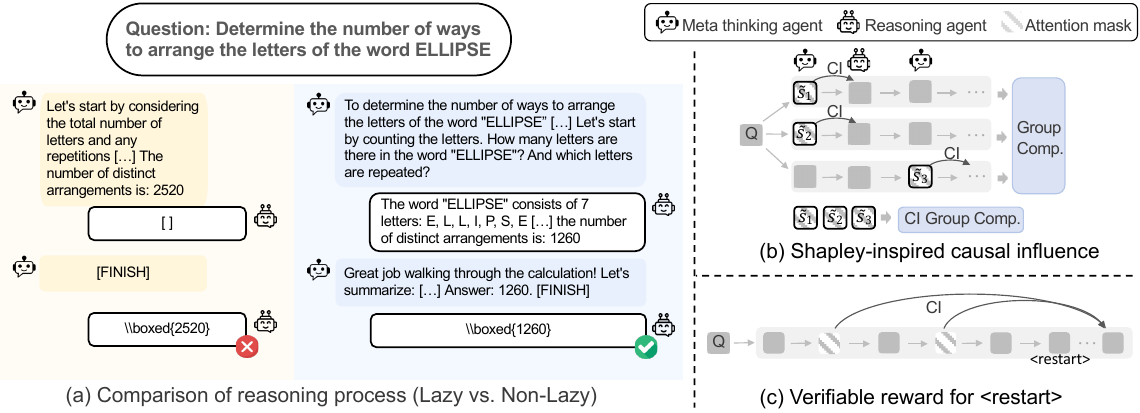}
    \caption{(a) Case study on lazy agents (full process in Appendix~\ref{appednix:casestudy}); (b–c) our proposed modules.}
    \label{fig:framework}
\end{figure}
In this section, we present empirical evidence of the lazy-agent problem in the multi-agent framework ReMA \citep{wan2025rema}. As illustrated in Fig.~\ref{fig:framework}(a), the reasoning agent often outputs blanks at intermediate steps, shifting the burden to the meta-thinking agent and ultimately leading to incorrect answers. By contrast, when both agents actively contribute, collaboration yields correct solutions.  
To quantify laziness, we measure the causal influence of an agent’s actions by adapting the attention-suppression method from \citet{bogdan2025thought}. Let $s_t$ denote an action taken by either agent. We suppress all attention (across layers and heads) to the tokens corresponding to $s_t$ and define the influence on the subsequent action $s_{t+1}$ as the KL divergence between the model’s logits with and without suppression. Intuitively, a small divergence indicates that the agent’s step has little impact on subsequent reasoning and thus reflects lazy behavior, whereas a large divergence shows that the step substantially shapes the reasoning process. See Appendix~\ref{sec:pre_causal} for experimental details.

% \zhiwei{[case study with figure and appendix]}

% To demonstrate the lazy agent issue is introduced by ReMA, we compare the causal influnce of agents with several settings: (1) initialized agents without training (2) agents trained with ReMA (3) agents trained with ReMA but with specifically designed prompt to avoid non-trivial response, where the prompt is detailed in Appendix \ref{}. The results are shown in Fig. \ref{fig:causal-effect-row} (a), (b) and (c), respectively.
% From the figures, we observe: \textbf{(1)} Compared to initialized version, the reasoning agent show significantly less causal contribution to the reasoning process compared to the meta-thinking agent, this indicates that the issue of lazy agent in ReMA framework. \textbf{(2)} With specifically design prompt to encourage more active and non-trivial conversations, the gap of causal effect between the reasoning agent and meta-thinking agent is smaller compared to original ReMA version, however, when compared to agents without any training, the reasoning agent still show less contribution compared to meta-thinking agent, this indicates that the reasoninga gent still learn to take short cut and do not learns to contribution as much as possible to the reasonign process, this motivate us to design a method that could mitigate the lazy agent issues in online reinforcement learning setting.
\begin{figure*}[h]
  \centering
  \begin{subfigure}[t]{0.251\textwidth}
    \includegraphics[width=\linewidth]{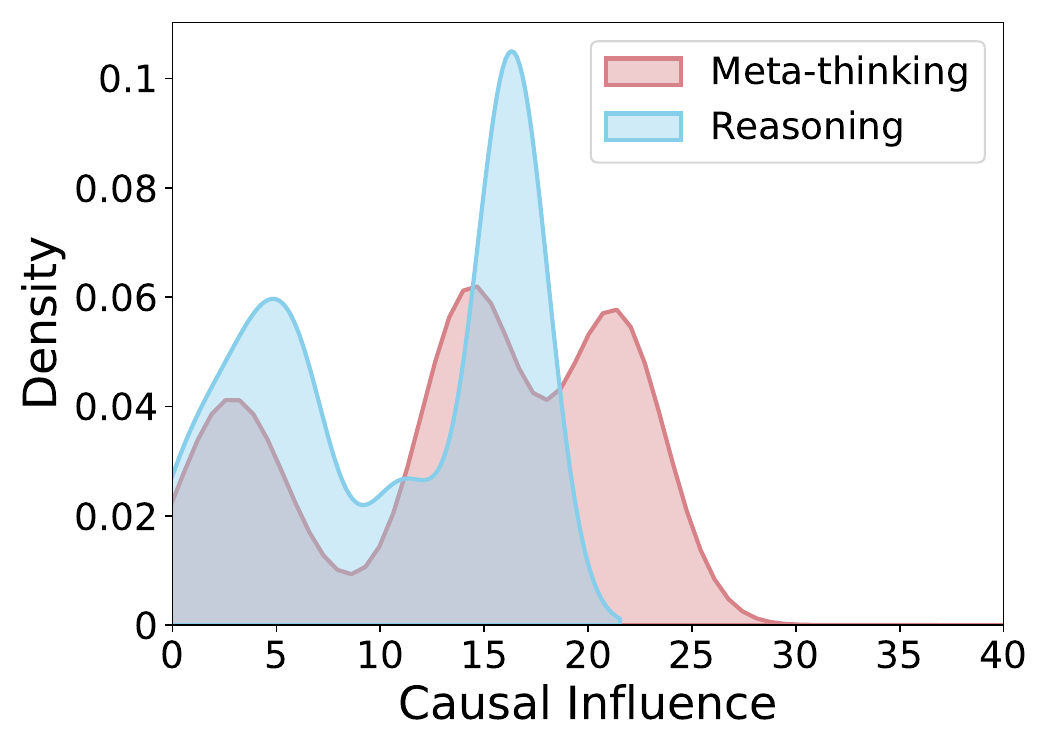}
    \caption{Initialized}
  \end{subfigure}
  \hspace{-6pt}
  \begin{subfigure}[t]{0.251\textwidth}
    \includegraphics[width=\linewidth]{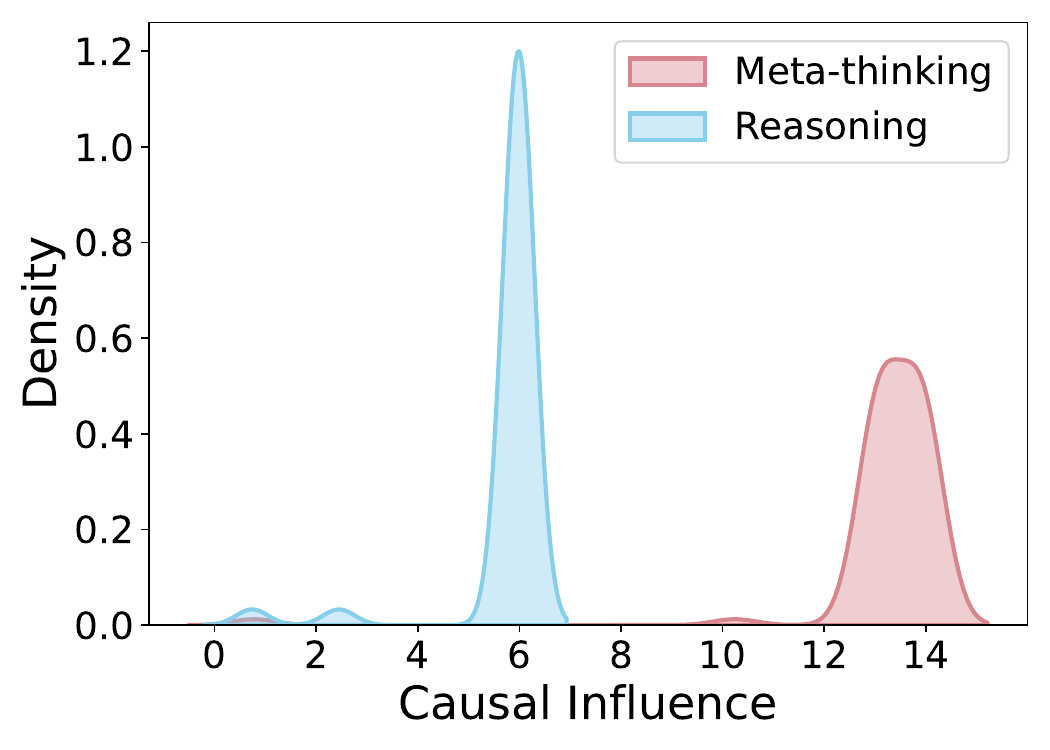}
    \caption{ReMA}
  \end{subfigure}
  \hspace{-6pt}
  \begin{subfigure}[t]{0.251\textwidth}
    \includegraphics[width=\linewidth]{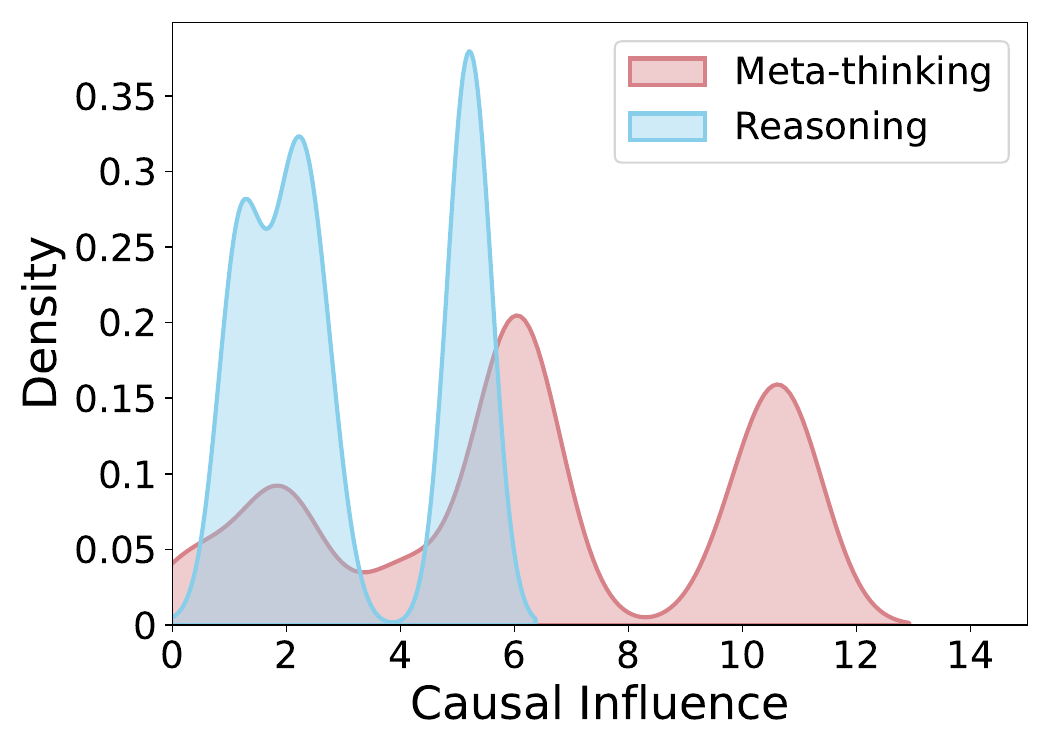}
    \caption{ReMA w/ prompt}
  \end{subfigure}
  \hspace{-6pt}
  \begin{subfigure}[t]{0.251\textwidth}
    \includegraphics[width=\linewidth]{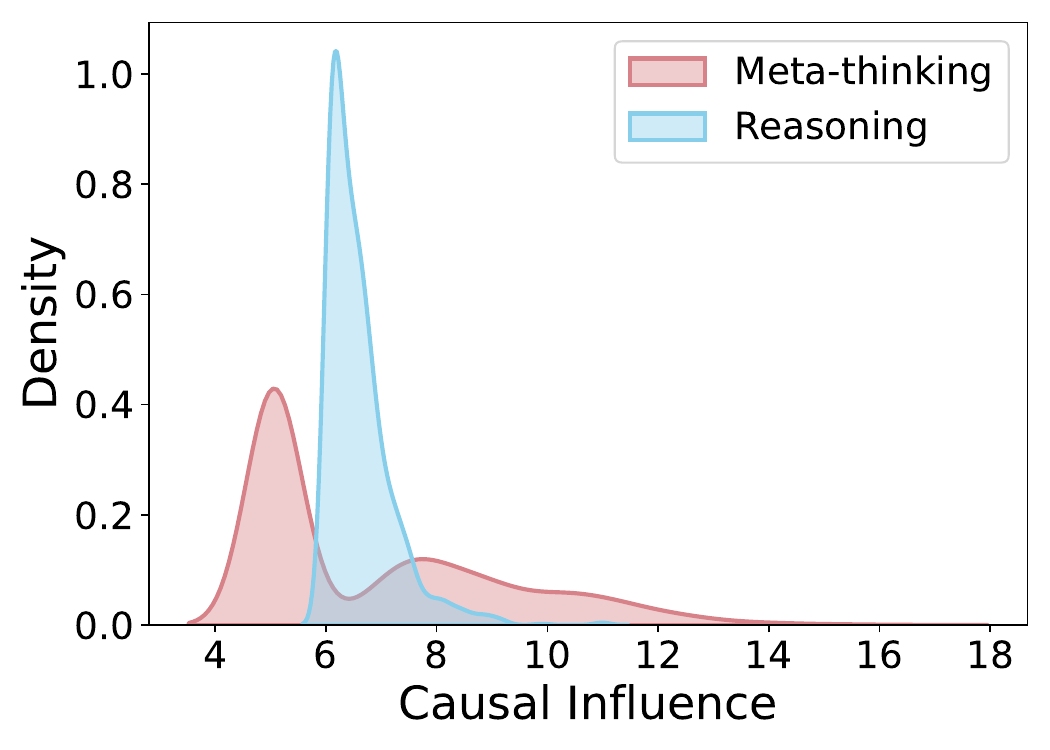}
    \caption{Ours}
  \end{subfigure}
  \vspace{-6pt}
  \caption{Causal effect comparison. Performance on MATH500 under different configurations: (a) 75.0, (b) 74.4, (c) 75.6, and (d) 78.4. }
  \label{fig:causal-effect-row}
\end{figure*}
To examine whether the lazy-agent issue arises from the ReMA framework, we compare causal influence under three settings: (1) untrained agents initialized from the base model, (2) agents trained with ReMA, and (3) agents trained with ReMA but prompted with instructions discouraging trivial responses (details in Appendix~\ref{sec:pre_causal}). The results are shown in Fig.~\ref{fig:causal-effect-row}(a–c).  
\textbf{(1)} Compared to the untrained baseline, the reasoning agent in the standard ReMA setting contributes substantially less than the meta-thinking agent, revealing clear lazy-agent behavior. This imbalance coincides with a performance drop from 75.0 to 74.4 on MATH500 despite training.  
\textbf{(2)} Adding a prompt to encourage non-trivial responses narrows the causal-effect gap and improves performance from 74.4 to 75.6. However, the reasoning agent still shows weaker influence than in the baseline, and the modest 0.6-point gain suggests reliance on shortcuts rather than meaningful reasoning.  
In summary, ReMA is prone to producing lazy agents. While prompt engineering can partially mitigate the issue, it does not fully resolve it. This underscores the need for more robust methods, particularly in online reinforcement learning, to ensure balanced agent contributions.

\section{Dr. MAMR: Multi-agent Meta-Reasoning Done Right}

% \zhiwei{maybe give a brief intro, from two perspectives}

% \suhang{what is the full name of Dr. MAMR?? Maybe need to give the full name in the introduction}

\subsection{Theoretical analysis on the emergence of lazy agent} 

Our preliminary experiments in Sec.~\ref{sec:preliminary} reveal the critical issue of lazy agents in advanced multi-agent reasoning frameworks. In this section, we provide a theoretical analysis of why such behavior emerges, even when the overall system performance appears to improve.  
Following GRPO~\citep{shao2024deepseekmath} and DAPO~\citep{yu2025dapo}, multi-turn GRPO~\citep{wan2025rema, wei2025webagent} introduces a normalization term to reduce sequence-level bias toward longer rollouts. As shown in Eq.~\ref{eq:rema}, the objective includes a factor \( \textcolor{red}{\tfrac{1}{T_i}} \) that averages the turn-level advantages across each trajectory.  
However, we find that this normalization introduces a structural bias:
\emph{given the same context, if two alternative actions produce trajectories with equal final reward but different numbers of turns, the model favors the action leading to fewer turns}.  
To illustrate, we consider the reasoning agent as an example and present the following theory:
% \xiaomin{Some example of phrasing: ``To understand the mechanism driving this lazy behavior, we analyze the structure of the multi-turn GRPO objective. We find it contains an inherent structural bias that, irrespective of the final reward, penalizes longer conversational trajectories. Theorem 1 formalizes this bias, quantifying the conditions under which the policy update will favor actions that lead to fewer turns." Also, replace or add to the saying ``shorter trajectories are favored" with ``the role with fewer, higher-impact turns learns faster"}
\begin{theorem}
Let
$
g_t(\tau) \;=\; \frac{1}{T(\tau)}\,Z_t(\tau)
$
be the gradient contribution at turn $t$ for trajectory $\tau$ with $Z_t(\tau) \;\triangleq\; \frac{1}{|y_t|} \sum_{j=1}^{|y_t|}
r_t(\theta)\,\hat{A}_{t,j}\,\nabla_\theta \log \pi_\theta\!\left(y_{t,j}\mid x,m_{\le t},y_{<t},y_{t,<j}\right).$  
Consider two continuations from the same prefix: a short trajectory 
$\tau^{\mathrm S}$ with horizon $T_{\mathrm S}$ and a long trajectory 
$\tau^{\mathrm L}$ with horizon $T_{\mathrm L} > T_{\mathrm S}$, leading to the same final reward.
Define
$
\kappa \;\triangleq\; 
\frac{\|Z_t(\tau^{\mathrm L})\|}
     {\|Z_t(\tau^{\mathrm S})\|}.
$
If $\kappa < \tfrac{T_{\mathrm L}}{T_{\mathrm S}}$, then 
$
\frac{\|g_t(\tau^{\mathrm S})\|}
     {\|g_t(\tau^{\mathrm L})\|} > 1.
$
\end{theorem}
This theorem shows that unless the aggregated contribution 
$Z_t(\tau^{\mathrm L})$ is at least $\tfrac{T_{\mathrm L}}{T_{\mathrm S}}$ times larger than 
$Z_t(\tau^{\mathrm S})$, the gradient update favors the trajectory with fewer turns. Importantly, this holds whether the advantages are both positive or both negative: in the latter case, although both trajectories are discouraged, the shorter one is penalized less.
Consequently, the model is biased toward actions that reduce the number of turns, even if longer trajectories are equally rewarding. 
Empirically, our results in Appendix~\ref{app:expforturn} show that reasoning processes with lazy-agent behavior (e.g., producing empty outputs or simply summarizing) consistently involve fewer turns than those without lazy agents at the initial training stages, which are critical in shaping policy behavior.
\textit{Together, our theorem and empirical findings explain the emergence of lazy-agent behavior}: the normalization bias steers optimization toward trajectories with fewer turns, and reasoning processes exhibiting lazy behavior naturally produce shorter trajectories, thereby receiving preferential reinforcement during training.
We emphasize that our analysis is distinct from Dr.GRPO \citep{liu2025understanding}, which examines token-level normalization. Their work removes length normalization to prevent the policy from favoring shorter correct answers or unnecessarily long incorrect ones. 
Furthermore, since the number of turns $T$ is far smaller than the number of tokens, the resulting bias in our setting is substantially more pronounced.

\subsection{Shapley-inspired causal influence}
To mitigate the optimization bias in multi-turn GRPO, we first remove the $\textcolor{red}{\tfrac{1}{T_i}}$ normalization in Eq.~\ref{eq:rema}, which alleviates but does not fully eliminate the lazy-agent issue as indicated in ablation study (Sec. \ref{exp:ablation}). Addressing this problem requires measuring the causal influence of each step during online training. In practice, however, the policy generates only a single continuation per step, so influence must be inferred from one trajectory. This creates two challenges: \textbf{(1)} it offers only a limited view of how a step shapes the reasoning process \citep{xu2025not}, and \textbf{(2)} it biases causal influence toward specific phrasings rather than underlying ideas \citep{pavlick2016simple, mccoy2019right, merrick2020explanation, li2024shapley}.  
Analogous to Shapley values \citep{li2021shapley}, which attribute contributions by averaging marginal effects across all coalitions, step-level causal influence in multi-agent RL should reflect average contributions across possible continuations rather than a single path. Directly computing such Shapley-style values, however, is infeasible because it would require extensive resampling during online RL. To make this tractable, we propose a stable Shapley-inspired causal influence measure.

We flatten each trajectory into a sequence \( s_{i,1}, s_{i,2}, \dots, s_{i,2T} \), where 
\( s_{i,2t-1} = m_{i,t} \) (meta-thinking) and \( s_{i,2t} = y_{i,t} \) (reasoning). 
Each step \( s_{i,t} \) is treated as an \emph{anchor step}, for which we form a group of semantically similar steps:
\[
G_S(s_{i,t}) \;=\; \{\, s_{j,t'} \mid s_{j,t'} \approx s_{i,t}, \; 1 \le j \le N,\; 1 \le t' \le 2T_j \,\},
\]
where $\approx$ denotes semantic similarity. Similarity can be easily measured through semantic distance (see Appendix \ref{app:implement}), ensuring that steps within the same group express a comparable idea. 
For each step \( s_{j,t'} \in G_S(s_{i,t}) \), we measure its \emph{one-step causal influence} on the next step. 
Let $h_{\le t'}$ be the local history up to and including step $t'$, and 
\( h_{\le t'}^{(j)\setminus t'} \) the masked history with $s_{j,t'}$ removed. 
We compare the probability of the next output under the full and masked histories:
\begin{equation}
\label{eq:stepci}
p_{\mathrm{full}}^{(j,t')} \;=\; \pi_\theta(s_{j,t'+1} \mid h_{\le t'}^{(j)}), 
\quad
p_{\mathrm{mask}}^{(j,t')} \;=\; \pi_\theta(s_{j,t'+1} \mid h_{\le t'}^{(j)\setminus t'}), 
\quad
\Delta\ell_{j,t'} \;\triangleq\; \log p_{\mathrm{mask}}^{(j,t')} - \log p_{\mathrm{full}}^{(j,t')}.
\end{equation}
Finally, the Shapley-inspired causal influence of an anchor step \( s_{i,t} \) is the average across its group:
\begin{equation}
\label{eq:ci}
\mathrm{CI}(s_{i,t}) \;=\; \frac{1}{|G_S(s_{i,t})|} \sum_{(j,t') \,:\, s_{j,t'} \in G_S(s_{i,t})} \Delta\ell_{j,t'}.
\end{equation}
Our method ensures reliable causal influence estimation by:  
\textbf{(1)} averaging the impact of semantically similar steps across rollouts to obtain a stable estimate of an idea’s overall contribution, and  
\textbf{(2)} aggregating different phrasings of the same idea to reduce wording bias in influence estimates.

\subsection{Reasoning agent deliberation for multi-Turn interactions}
\label{sec:deliberation}

As each agent contributes more actively, the number of dialogue turns between the meta-thinking and reasoning agents increases. However, prior work shows that longer multi-turn interactions can degrade performance: \citet{laban2025llms} compare LLMs in (1) a single-turn setting where the full task is given in one prompt, and (2) a multi-turn setting where the task is decomposed into incremental prompts. They report consistent performance drops in the multi-turn condition, likely because LLMs overcommit to underspecified early context and struggle to recover from initial errors.  
These findings imply that while multi-agent collaboration can enrich reasoning, it also heightens vulnerability to error propagation when intermediate turns introduce ambiguity. If we view the meta-thinking agent as a user providing step-by-step instructions, then more interactions risk the reasoning agent becoming “lost” in dialogue, as observed by \citet{laban2025llms}. 
To mitigate this, we hypothesize that allowing the reasoning agent to discard prior responses, aggregate the meta-thinking prompts, and restart reasoning would be beneficial.

\begin{figure}[t]
    \centering
    \includegraphics[width=0.65\linewidth]{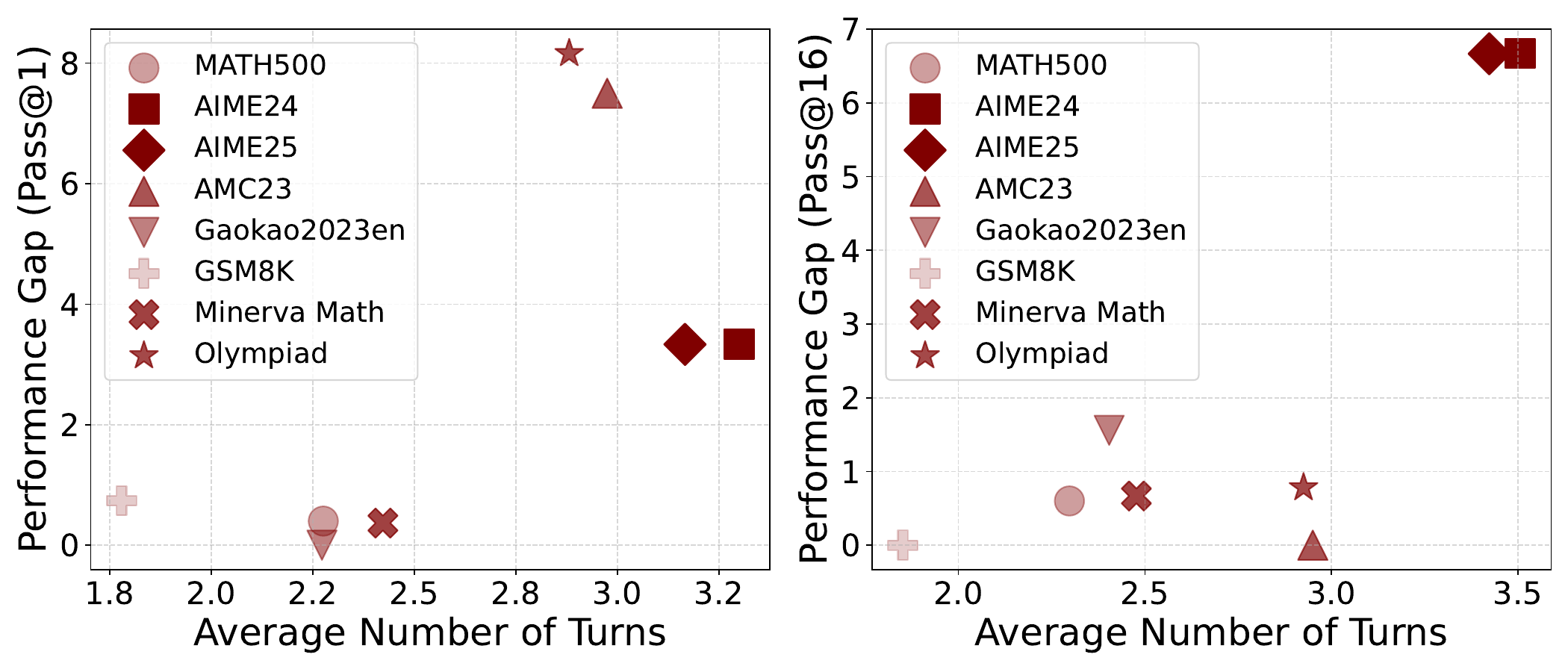}
    \caption{
        Performance gap between ReMA+ and ReMA across 8 benchmarks.
        Left: Pass@1. 
        Right: Pass@16. Darker colors represent more difficult benchmarks.
    }
    \label{fig:pre_turn}
\end{figure}
To validate this assumption, we conduct preliminary experiments following the ReMA framework to obtain a meta-thinking agent and a reasoning agent. We then compare the standard ReMA framework with a modified variant, \textbf{ReMA+}, where the reasoning agent is guided by a refined system prompt that enables it to adaptively discard its previous outputs when necessary. The full prompt design and additional details are provided in Appendix~\ref{sec:pre_deliberation_appendix}.  
In this experiment, we modify the system prompt only at inference time. Effectiveness is evaluated by (i) the performance gap in validation accuracy between ReMA+ and ReMA. Results across eight benchmarks are shown in Fig.~\ref{fig:pre_turn}.  
From the figure, we observe:  
\textbf{(1)} Without explicit training for deliberation, ReMA+ consistently matches or outperforms ReMA, with gains of about 8\% on AMC23 and Olympiad under Pass@1, and 7\% on AIME24 and AIME25 under Pass@16.  
\textbf{(2)} Even under Pass@16, which gives both frameworks ample chances to succeed, ReMA+ outperforms ReMA on 6 of 8 benchmarks—demonstrating that while LLMs have this capacity, explicit prompting is required for consistent behavior.  
\textbf{(3)} The performance gap widens as benchmark difficulty and the number of turns increase, highlighting the value of adaptive deliberation in extended multi-agent interactions.

Building on these observations, we design a method that trains the reasoning agent to adaptively discard its previous outputs to improve the likelihood of producing a correct final answer. Specifically, we introduce a control token, \texttt{<restart>}, which instructs the agent to discard prior responses, consolidate the instructions, and begin a fresh attempt. To assess the benefit of this mechanism, we develop a novel verifiable criterion that measures how discarding history affects the model’s probability of generating the correct final output.

\textbf{Verifiable Reward for Restart.}  
Consider the $i$-th rollout where the reasoning agent outputs \texttt{<restart>} at turn $t$. In this case, we mask all reasoning-agent outputs that occur strictly before $t$:  
$
\mathcal{Y}_{<2t}^{(i)} \;\triangleq\; \{ s_{j,k} \mid k < 2t,\; k \bmod 2 = 0 \},
$
and define the causal influence of the restart action on the final reasoning step \( \mathbf{y}_T^{(i)} = s_{i,2T} \) as
\begin{equation}
\label{eq:ci-restart}
\Delta\ell_{i,t}
\;\triangleq\; 
\log \pi_\theta(s_{i,2T} \mid h_{\leq 2T}^{(i) \setminus \mathcal{Y}_{<2t}^{(i)}})
\;-\;
\log \pi_\theta(s_{i,2T} \mid h_{\le 2T}^{(i)}).
\end{equation}
We define a binary outcome reward $z_i$ as $+1$ if the final answer \( \mathbf{y}_T^{(i)} \) is correct, and $-1$ if it is incorrect. The restart reward is then
\begin{equation}
\label{eq:restart}
r^{\text{restart}}_{i,t} \;=\;
\begin{cases}
+1, & \text{if } (z_i=+1 \,\wedge\, \Delta\ell_{i,t}>0) \;\;\text{or}\;\; (z_i=-1 \,\wedge\, \Delta\ell_{i,t}<0), \\[6pt]
-1, & \text{if } (z_i=+1 \,\wedge\, \Delta\ell_{i,t}<0) \;\;\text{or}\;\; (z_i=-1 \,\wedge\, \Delta\ell_{i,t}>0), \\[6pt]
0,  & \text{if } \Delta\ell_{i,t}=0.
\end{cases}
\end{equation}
This reward provides a verifiable signal of whether the restart improves or worsens the model’s belief in the final output. If the final answer is correct ($z_i=+1$), the restart is rewarded when masking prior reasoning increases confidence (\( \Delta\ell_{i,t} > 0 \)); otherwise it is penalized. The converse holds when the final answer is incorrect ($z_i=-1$).

\textbf{Aggregated Step-Level Advantage.}  
Let \( \mathrm{CI}(s_{i,t}) \) (Eq.~\ref{eq:ci}) denote the Shapley-inspired causal influence for step \(t\) in rollout \(i\), and let \( r^{\mathrm{restart}}_{i,t} \) be the verifiable restart reward (Eq.~\ref{eq:restart}), defaulting to \(0\) if no \texttt{<restart>} is issued.  
For normalization, each signal \(x\) is first rescaled to \(\widetilde{x} \in [-1,1]\) using min--max scaling (details in Appendix~\ref{app:implement}), and then standardized across all rollouts with mean \(\mu_{\widetilde{X}}\) and standard deviation \(\sigma_{\widetilde{X}}\) by 
$
\mathcal{Z}_X(x) \;\triangleq\; \frac{\widetilde{x} - \mu_{\widetilde{X}}}{\sigma_{\widetilde{X}}}.
$
We apply this procedure to obtain the normalized causal signal \(\tilde{C}_{i,t} \;=\; \mathcal{Z}_{\{\mathrm{CI}\}}\bigl(\mathrm{CI}(s_{i,t})\bigr)\) and restart signal \(\tilde{R}_{i,t} \;=\; \mathcal{Z}_{\{r^{\mathrm{restart}}\}}\bigl(r^{\mathrm{restart}}_{i,t}\bigr)\).  
The overall step-level advantage is then defined as a weighted combination:
\begin{equation}
\label{eq:final}
A^{\mathrm{step}}_{i,t} \;=\; \tilde{A}_{i,t} \,+\, \alpha\,\tilde{C}_{i,t} \,+\, \beta\,\tilde{R}_{i,t},
\end{equation}
where \( \tilde{A}_{i,t} \) is the normalized outcome-based advantage, and \( \alpha, \beta \) are tunable hyperparameters. The training objective for Dr. MAMR builds on Eq.~\ref{eq:rema}, removing the $\tfrac{1}{T}$ normalization and replacing the advantage function with Eq.~\ref{eq:final}. See implementation details in Appendix \ref{app:implement}.

% \textbf{Aggregated Step-Level Advantage.}
% Let \( \mathrm{CI}(s_{i,t}) \) (from Eq.~\ref{eq:ci}) denote the group-averaged causal influence for step \( t \) in rollout \( i \), and let \( r^{\mathrm{restart}}_{i,t} \) be the verifiable restart reward (Eq.~\ref{eq:restart}), defaulting to \( 0 \) if no \texttt{<restart>} is issued.
% For a batch of \( |G| \) rollouts, we compute a normalized trajectory-level advantage by first mapping values to the bounded range \([-1,1]\) via min--max scaling and then applying standardization to the rescaled values:
% \begin{equation}
% \mathcal{Z}_X(x) \;\triangleq\; \frac{\widetilde{x} - \mu_{\widetilde{X}}}{\sigma_{\widetilde{X}}},\quad \widetilde{x} \;=\; 2 \cdot \frac{x - \min(X)}{\max(X) - \min(X)} - 1,
% \end{equation}
% where \(\mu_{\widetilde{X}}\) and \(\sigma_{\widetilde{X}}\) denote the mean and standard deviation of the min--max normalized set \(\widetilde{X}\), computed over all relevant steps across the rollouts.
% Define the normalized causal and restart signals:
% \[
% \tilde{C}_{i,t} \;=\; \mathcal{Z}_{\{\mathrm{CI}\}}\bigl(\mathrm{CI}(s_{i,t})\bigr),
% \qquad
% \tilde{R}_{i,t} \;=\; \mathcal{Z}_{\{r^{\mathrm{restart}}\}}\bigl(r^{\mathrm{restart}}_{i,t}\bigr).
% \]

% We then define the overall turn-level advantage as a weighted combination:
% \begin{equation}
% A^{\mathrm{step}}_{i,t} \;=\; \tilde{A}_{i,t} \,+\, \alpha\,\tilde{C}_{i,t} \,+\, \beta\,\tilde{R}_{i,t},
% \end{equation}
% where \( \tilde{A}_{i,t} \) is the normalized outcome-based advantage, and \( \alpha, \beta\) are tunable hyperparameters.

\section{Experiments}

\subsection{Experiment Settings}
\label{sec:expsetting}

\textbf{Dataset and Benchmarks:} We conduct experiments on mathematical reasoning by training models on DeepScaleR dataset \citep{deepscaler2025}. The optimized agents are then evaluated across seven benchmarks: MATH500 \citep{lightman2023let}, GSM8K \citep{cobbe2021training}, AIME 
% \footnote{\texttt{https://huggingface.co/datasets/AI-MO/aimo-validation-aime}}
, AMC23 \citep{numina_math_datasets}
% \footnote{\texttt{https://huggingface.co/datasets/AI-MO/aimo-validation-amc}}
, GaoKao2023En \citep{zhang2023evaluating}, Minerva Math \citep{lewkowycz2022solving}, and OlympiadBench \citep{he2024olympiadbench}. See implementation details in Appendix \ref{app:implement}.

\textbf{Baselines and Models:} 
We compare Dr.MAMR against three baselines: (1) \textbf{GRPO} \citep{guo2025deepseek}, where the base model is trained with vanilla GRPO in a single-agent setting; (2) \textbf{VRP (CoT)} \citep{wan2025rema}, where the base model is prompted step by step to operate within the meta-thinking and reasoning framework; and (3) \textbf{ReMA} \citep{wan2025rema}, a multi-agent meta-reasoning framework trained with multi-turn GRPO.
We conduct training and evaluation on the Qwen2.5 family, using the 3B, 7B, and 14B Instruct models \citep{qwen2.5}.

\subsection{Results on Seven Benchmarks}

\paragraph{Question 1:}\textit{How does Dr. MAMR perform in reasoning tasks compared to baselines?}

Table~\ref{tab:main} reports pass@1 performance across seven benchmarks using 7B and 14B base models. (See Appendix \ref{app:mainfor3b} for 3B results). 
From the results, we observe the following:
\textbf{(1)} ReMA consistently underperforms compared to single-agent GRPO, highlighting the severity of the lazy agent issue and its detrimental effect on multi-agent performance.
\textbf{(2)} Dr. MAMR consistently outperforms single-agent GRPO across all base models. Notably, the performance gain increases with larger base models that exhibit stronger instruction-following capabilities. This suggests that mitigating the lazy agent problem enables effective collaboration between agents and leads to better outcomes. This also points to a promising direction: improving the instruction-following ability of base models, which supports more effective communication and collaboration between agents, can in turn lead to better overall system performance.
\textbf{(3)} \textbf{Our Dr. MAMR elevates the multi-agent system baseline from performing worse than single-agent GRPO to clearly outperforming it}, demonstrating the potential of multi-agent frameworks in solving complex reasoning tasks when carefully designed.
\begin{table}[h]
\centering
\caption{Performance on math benchmarks.}
\label{tab:main}
\begin{tabular}{c|c|ccccc}
\toprule
\textbf{Model} & \textbf{Benchmark} & \textbf{GRPO} & \textbf{VRP (CoT)} & \textbf{ReMA} & \textbf{Dr. MAMR (Ours)} \\
% \midrule
% \multirow{8}{*}{\shortstack[l]{Qwen2.5\\-3B\\-Instruct}}
% &MATH500          & 65.60 & 65.20  & 62.60 & 66.20\\
% &GSM8K            & 85.30 & 72.02  & 83.17 & 85.37\\
% &AIME24           & 13.33 & 3.33   & 3.33  & 16.67\\
% &AMC23            & 40.00 & 20.00  & 42.50 & 50.00\\
% &Gaokao2023en     & 54.30 & 30.91  & 52.73 & 55.33\\
% &Minerva Math     & 31.20 & 16.91  & 26.47 & 32.35\\
% &Olympiad Bench   & 30.20 & 6.07   & 27.56 & 30.57\\
% \cmidrule(lr){2-6}
% &\textbf{Average} &    45.70   &   30.63     &  42.62     & 48.07\\
\midrule
\multirow{8}{*}{\shortstack[l]{Qwen2.5\\-7B\\-Instruct}} 
&MATH500          & 75.50  & 75.00  & 74.40 & \textbf{78.60} \\
&GSM8K            & 90.50  & 92.04  & 90.60 & \textbf{92.12} \\
&AIME24           & 16.67  & 6.67   & 13.33 & \textbf{20.00} \\
&AMC23            & 55.00  & 47.50  & 50.00 & \textbf{62.50} \\
&Gaokao2023en     & 64.60  & 56.62  & 57.92 & \textbf{65.20} \\
&Minerva Math     & 34.70  & 35.66  & 34.93 & \textbf{38.24} \\
&Olympiad Bench   & 48.60  & 38.22  & 42.58 & \textbf{52.34} \\
\cmidrule(lr){2-6}
&\textbf{Average} & 55.08 & 50.24 & 51.97 & \textbf{58.43} \\
\midrule
\multirow{8}{*}{\shortstack[l]{Qwen2.5\\-14B\\-Instruct}}
&MATH500          & \textbf{80.60}  & 78.40  &   79.20    & 80.40 \\
&GSM8K            & \textbf{94.50}  & 92.87  &   93.59    & 93.69 \\
&AIME24           & 16.67  & 10.00  &   13.33    & \textbf{26.67} \\
&AMC23            & 60.00  & 55.00  &   60.00    & \textbf{67.50} \\
&Gaokao2023en     & 64.90  & 66.23  &   67.53    & \textbf{69.09} \\
&Minerva Math     & 41.50  & 38.60  &   41.91    & \textbf{43.02}\\
&Olympiad Bench   & 48.20  & 46.78  &  45.12     &\textbf{57.03} \\
\cmidrule(lr){2-6}
&\textbf{Average} &  58.05      &   55.41     &  57.24     & \textbf{62.49} \\
\bottomrule
\end{tabular}
\end{table}
\subsection{Training curves}
\paragraph{Question 2:} \textit{How does the causal influence of agents evolve during training?}

\begin{figure*}
  \centering
  \begin{subfigure}[t]{0.45\textwidth}
    \includegraphics[width=\linewidth]{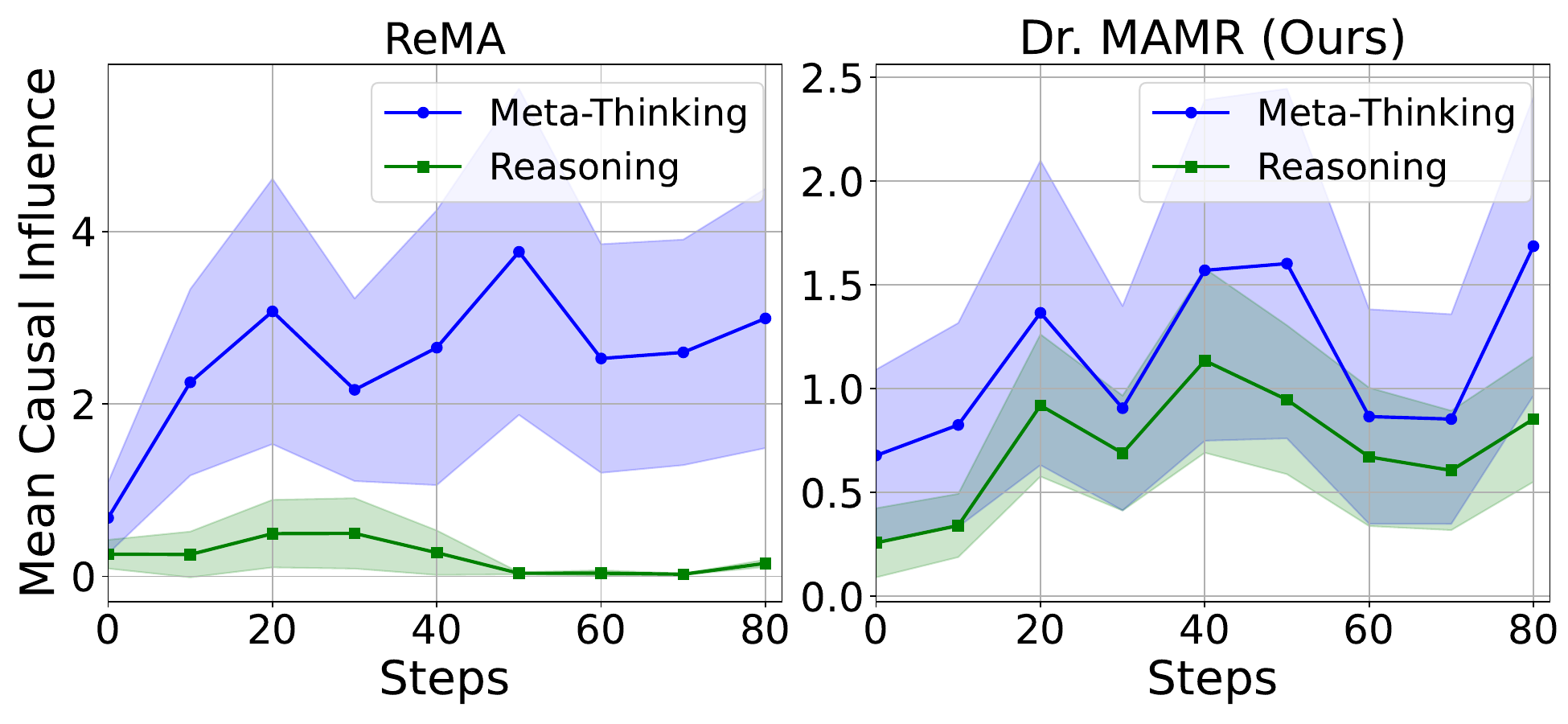}
    \caption{Causal influence across training steps.}
  \end{subfigure}
  \hspace{-6pt}
  \begin{subfigure}[t]{0.261\textwidth}
    \includegraphics[width=\linewidth]{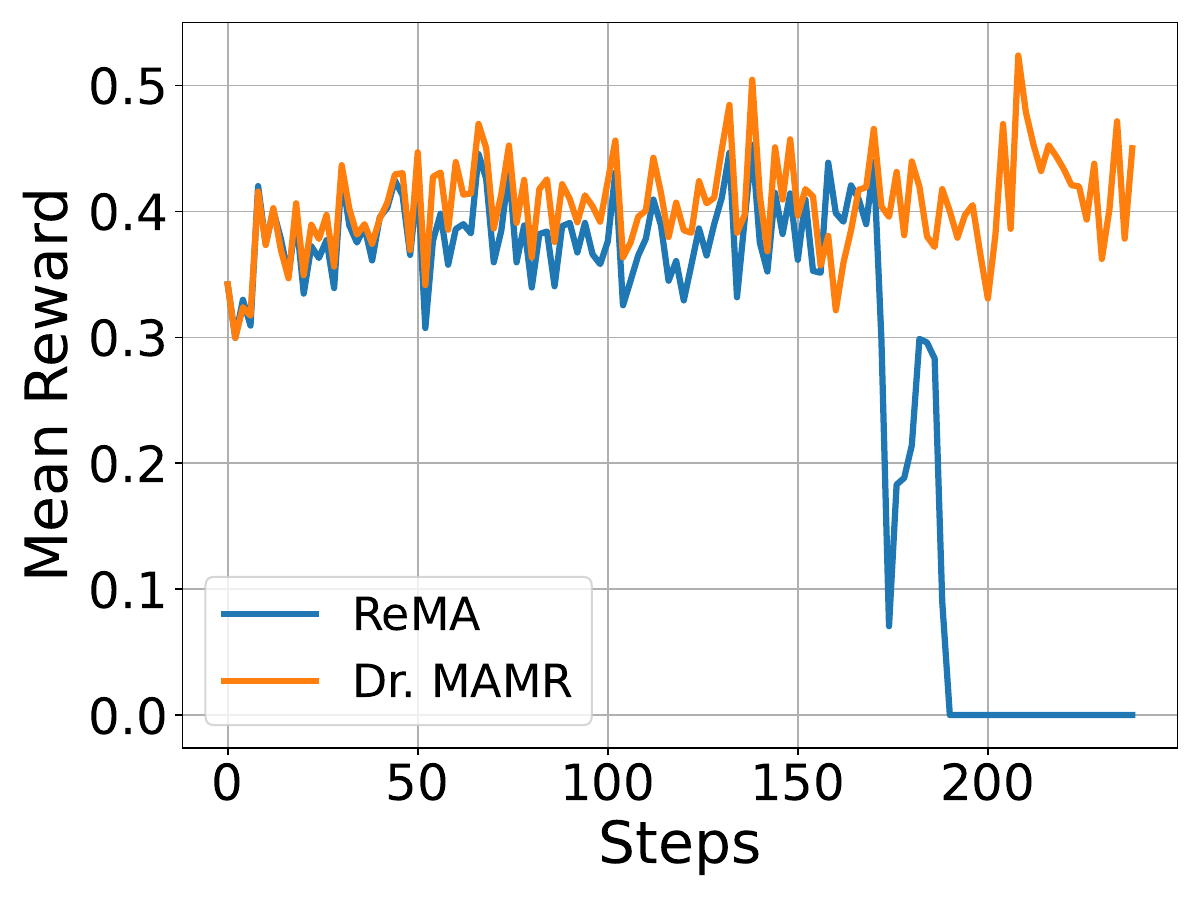}
    \caption{Training Curve.}
  \end{subfigure}
  \begin{subfigure}[t]{0.278\textwidth}
      \includegraphics[width=\linewidth]{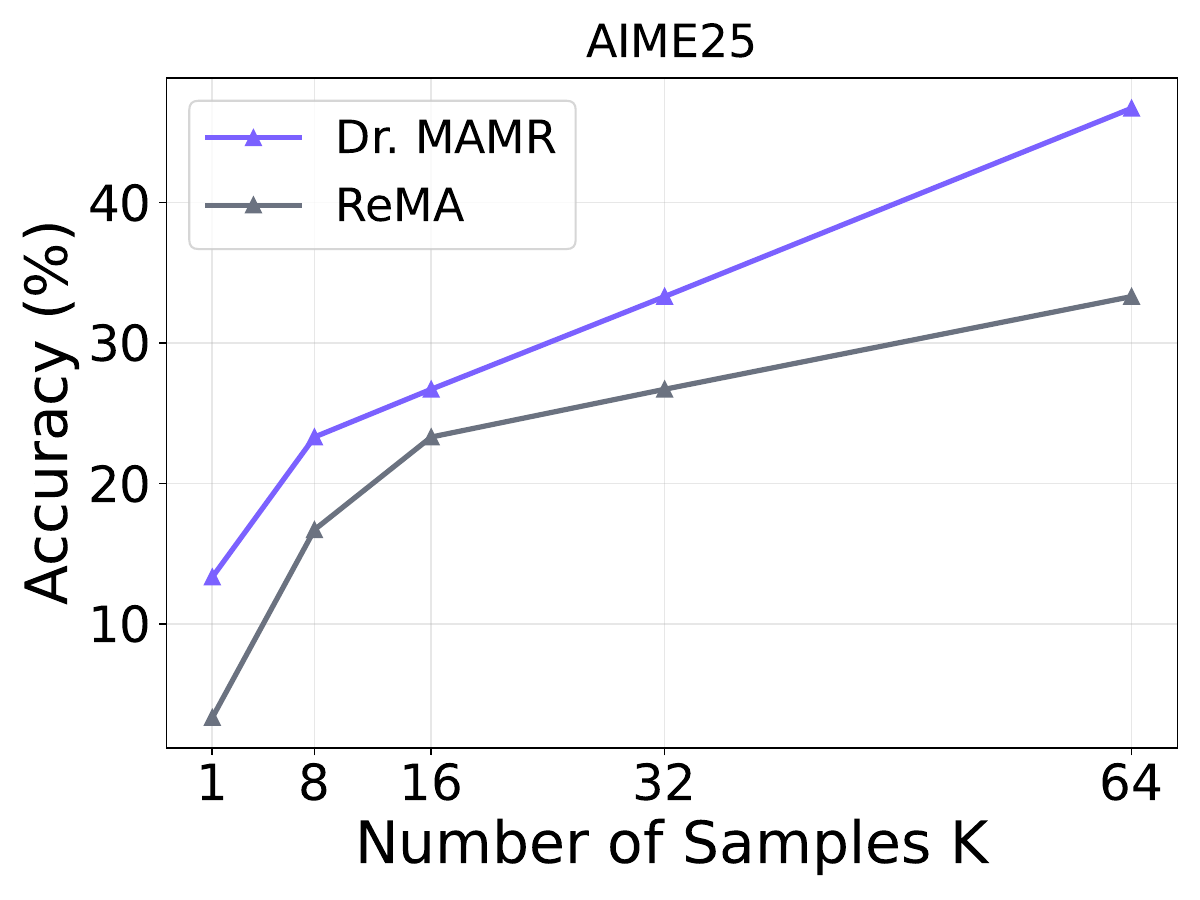}
    \caption{Pass@$K$ Performance.}
  \end{subfigure}
  \caption{Results on causal influence, training stability, and pass@K.}
  \label{fig:combinedexp}
\end{figure*}

In this section, we present a case study on the 7B model, examining how the causal influence of the meta-thinking agent and the reasoning agent evolves during training under our Dr. MAMR framework, compared to ReMA. We report the results in Fig.~\ref{fig:combinedexp} (a). From the figure, we observe: \textbf{(1)} Under ReMA, the reasoning agent’s causal influence initially increases slightly at the beginning of training but then steadily decreases, eventually approaching zero, while the meta-thinking agent’s influence grows significantly as it comes to dominate the reasoning process. This indicates that naive multi-turn GRPO risks collapsing the system into an effective single-agent setup, losing the benefits of collaboration. \textbf{(2)} In contrast, under Dr. MAMR, the reasoning agent’s influence steadily increases, while the meta-thinking agent also grow consistently, indicating that both agents contribute meaningfully. This balanced collaboration explains why Dr. MAMR achieves superior performance across diverse reasoning tasks compared to ReMA.

% \begin{figure}
%     \centering
%     \includegraphics[width=0.5\linewidth]{figs/causal_influence.pdf}
%     \caption{Caption}
%     \label{fig:causal}
% \end{figure}

% \subsection{Training curves}
\paragraph{Question 3:} \textit{How does Dr. MAMR stabilize multi-agent RL compared to the baseline?}

Training stability is a key challenge in multi-agent RL. Thus, we compare the training curves and report mean rewards of Dr. MAMR and ReMA on the training data, with results for the 7B model shown in Fig.~\ref{fig:combinedexp}(b) and additional 3B results provided in Appendix~\ref{app:3Bcurve}. From the figure, we observe that after 50 steps, Dr. MAMR achieves clearly superior performance. After 150 steps, ReMA collapses with its reward dropping to zero, whereas Dr. MAMR maintains stable training throughout. This demonstrates the benefit of addressing the lazy-agent issue for stabilizing multi-agent RL.

\subsection{Scaling on Pass@$K$}
\paragraph{Question 4.} \textit{How does Dr. MAMR perform when scaling to pass@K?}

We examine test-time scaling \citep{muennighoff2025s1, zhang2025survey} by comparing the pass@$K$ performance of Dr. MAMR and ReMA, which measures whether the correct solution appears within the best result of $K$ independent attempts. Results on the most challenging benchmark, AIME25, are presented in Fig.~\ref{fig:combinedexp}(c), with additional benchmarks reported in Appendix~\ref{app:scaling}. The figure shows that \textbf{the performance gap between Dr. MAMR and ReMA widens as $K$ increases, highlighting Dr. MAMR’s strength in tackling difficult tasks}.

\subsection{Ablation Study}
\label{exp:ablation}
\paragraph{Question 5.} \textit{How does each component contribute to reasoning performance?}

We compare the full model against three variants:
(1) w/o Normalization Debias (w/o ND), which retains the turn-level normalization from ReMA;
(2) w/o Shapley-inspired causal influence (w/o CI); and
(3) w/o Restart Behavior (w/o RB).
We train these variants and report their benchmark performance in Table \ref{tab:ablation}.
From the table, we observe:  
\textbf{(1)}  
Dropping either normalization or causal influence causes clear performance drops, showing their complementary role in discouraging shortcuts and promoting balanced contributions.  
\textbf{(2)}  
Removing restart behavior also degrades performance across benchmarks, though less severely, highlighting its value in enabling recovery from mistakes and sustaining stable reasoning.
We provide a case study on restart behavior in Appendix \ref{app:caserestart}.
\begin{table}[h]
    \centering
    \caption{Ablation study on the 7B model.}
    \label{tab:ablation}
    \begin{tabular}{c|cccc}
    \toprule
       Variant  & AIME24 & AMC23 & Gaokao2023en & Olympiad Bench\\
    \midrule
       Dr.MAMR  & 20.00 & 62.50 & 65.20 & 52.34 \\
       w/o ND   & 13.33 &  55.00 &  63.64 &47.85\\
       w/o CI   & 13.33 &  52.50 & 63.38   &45.31\\
       w/o RB   & 16.67 &  57.50 & 63.90  & 50.58\\
    \bottomrule
    \end{tabular}
    \label{tab:placeholder}
\end{table}
\section{Conclusion}
We identify the issue of lazy agents in multi-agent LLM reasoning and trace the issue to the loss structure of multi-turn GRPO. To address it, we introduce a Shapley-inspired causal influence measure and a verifiable reward for restart behavior. Experiments across diverse benchmarks demonstrate that Dr.~MAMR effectively mitigates lazy behavior and surpasses strong single- and multi-agent baselines, unlocking the potential of multi-agent frameworks for complex reasoning tasks.

% \newpage

% \section*{Ethics Statement}
% This work does not involve human subjects or the collection of new datasets. Experiments use established corpora and benchmarks
% under their licenses.
% \section*{Reproducibility Statement}
% We provided a detailed description implementation details in Appendix \ref{app:implement} and experimental settings in Appendix \ref{app:expdetail}. Our code is available here: \href{https://anonymous.4open.science/r/MAMR-D666/}{https://anonymous.4open.science/r/MAMR-D666/}.

\bibliography{iclr2026_conference}
\bibliographystyle{iclr2026_conference}

\newpage

\appendix

\section{Detailed Related Works}
\label{app:relatedwork}

\subsection{Multi-agent RL}
Multi-agent RL addresses how multiple agents coordinate in a shared environment to maximize collective performance, with a central challenge being credit assignment: determining each agent’s contribution to the overall reward. Classical solutions include value decomposition (VDN; \citealp{sunehag2018value}), counterfactual baselines (COMA; \citealp{foerster2018counterfactual}), regression-based reward functions with default-action substitution (Dr.Reinforce; \citealp{castellini2022difference}), social influence estimation via KL divergence \citep{jaques2019social}, role-based coordination through role networks \citep{wang2020roma}, and model-based transition prediction to measure influence \citep{liu2023lazy}. With the advent of large language model (LLM) agents, MARL techniques have been extended to multi-turn reasoning and cooperative dialogue: turn-level credit assignment to reduce misattributed rewards \citep{zeng2025reinforcing}, SWEET-RL for critic-driven step-wise rewards \citep{zhou2025sweet}, MARFT for alleviating agent inactivity and communication inefficiency \citep{liao2025marft}, RAGEN for addressing the “Echo Trap” caused by coarse reward signals \citep{wang2025ragen}, and MAGRPO for framing LLM collaboration as cooperative MARL with tailored reward design \citep{liu2025llm}. A persistent challenge across both classical and LLM-based MARL is the emergence of lazy agents that contribute little while relying on others. Recent work therefore explores causal influence estimation \citep{bogdan2025thought, nguyen2025causalplan, liu2024large}, introducing methods such as black-box resampling and attention suppression to quantify how an agent’s utterance shapes subsequent decisions, thereby enabling finer-grained credit assignment and mitigating agent inactivity.

\subsection{Hierarchical RL}
\label{app:relatedwork-hrl}
Hierarchical multi-agent systems coordinate cooperation by assigning high-level controllers to decompose tasks for lower-level workers, a design shown to improve scalability, robustness, and long-horizon reasoning. \citep{yu2024fincon} introduce a manager–analyst paradigm for structured decomposition, while Chain-of-Agents (CoA) leverages chained communication to approximate hierarchical coordination with greater flexibility \citep{zhang2024chain}. Empirical studies further demonstrate that boss–worker hierarchies outperform flat or linear structures under failure conditions \citep{huang2024resilience}. Memory-oriented approaches, exemplified by Tracing Hierarchical Memory for Multi-Agent Systems, highlight how layered storage and retrieval mechanisms enable adaptive long-horizon collaboration \citep{zhang2025g}. On the planning side, Hierarchical Planning for Complex Tasks with Knowledge-Graph RAG and Symbolic Verification integrates structured decomposition with formal verification \citep{cornelio2025hierarchical}. More explicit architectures, such as HMAW and ReAcTree, explore CEO–Manager–Worker hierarchies or adaptive tree structures for general task allocation \citep{liu2024towards, choi2025reactree}.
Hierarchical reinforcement learning (HRL) organizes control as high-level planning over subgoals with low-level execution, enabling agents to operate over long horizons via temporal abstraction and reusable skills. Early neural HRL emphasized top-down goal setting—e.g., high-level goal embeddings steering a low-level policy \citep{vezhnevets2017feudal} and quickly expanded toward unsupervised skill discovery to populate the low-level option set \citep{bagaria2019option}. Beyond purely reactive control, symbolic reasoning has been fused with HRL to support plan construction and revision \citep{yamamoto2018hierarchical}. Recent advances improve the reliability and interpretability of options themselves: programmatic, human-readable sub-policies selected by the high-level planner enhance generalization to longer tasks \citep{lin2024hierarchical}, while theoretically grounded supervision clarifies how limited human feedback can be efficiently allocated across hierarchical levels in goal-conditioned settings \citep{yan2024theoretical}. Complementary lines strengthen the low level: disentangling unsupervised skill discovery yields cleaner building blocks \citep{hu2024disentangled}, and integrating imperfect expert priors improves multi-agent coordination under hierarchical control \citep{liu2024integrating}.

\subsection{LLM Reasoning}
Large Language Models (LLMs) have demonstrated strong performance across a wide range of natural language tasks \citep{li2024rule, wang2024comprehensive, brown2020language, chowdhery2023palm, du2022glam, dubey2024llama, wang2025agenttts, wenzek2019ccnet, zhangcatastrophic, li2025thinking, li2024rule}. Early research found that prompting models to reason step by step, an approach known as chain-of-thought (CoT) prompting, can significantly improve performance on arithmetic, commonsense, and symbolic reasoning tasks by eliciting intermediate reasoning steps \citep{wei2022chain, kojima2022large, nye2021show}. Building on this, researchers have explored non-linear reasoning structures. For example, Tree-of-Thoughts (ToT) organizes candidate reasoning paths into a search tree with lookahead capabilities, while Graph-of-Thoughts (GoT) generalizes reasoning to arbitrary graphs of “thought” nodes and edges, expanding the space for structured deliberation \citep{yao2023tree, besta2024graph}. These advances have inspired the development of Large Reasoning Models (LRMs), models explicitly trained for multi-step reasoning \citep{guo2025deepseek, achiam2023gpt, grattafiori2024llama, xu2023re, zhou2022least, wu2024comparative, qi2024mutual, chae2024language}. Typically, LRMs undergo supervised fine-tuning followed by a reinforcement learning stage, and have achieved state-of-the-art results on challenging tasks such as math, coding, and task planning \citep{jaech2024openai, guo2025deepseek, comanici2025gemini, yang2024qwen2, yang2025qwen3, lightman2023let, wang2023math}. The success of strong single-model reasoners has also spurred multi-agent approaches, where complex tasks are decomposed and coordinated among specialized LLM agents via role assignment, orchestration, and debate, mirroring human teamwork \citep{li2023camel,wu2024autogen,chen2023autoagents,du2023improving, yuanreinforce}.
% Methodologically, HRL has addressed distribution shift between levels and the credit-assignment gap using off-policy corrections and hierarchical interfaces \citep{nachum2018data}, and explored information-theoretic objectives to learn stable subgoals \citep{osa2019hierarchical}. The LLM extends HRL by letting language models act as high-level planners that decompose tasks into executable steps for RL controllers—offline hierarchical pipelines show natural-language plans converted to subgoals for low-level policies \citep{hu2025divide}, and hybrid neuro-symbolic planners combine knowledge-graph retrieval and formal checks to validate subgoals before execution \cite{cornelio2025hierarchical}. Multi-agent formulations make the hierarchy explicit within LLM systems themselves: a meta-thinking agent plans and adapts subgoals while a reasoning agent executes in alternating turns, enabling end-to-end reinforcement learning for complex reasoning \citep{wan2025rema}. Classic hierarchical architectures remain foundational—scalable off-policy goal interfaces \citep{nachum2018data}, compositional option sets \citep{barreto2019option}, and learned hierarchical abstractions citep{zhang2021hierarchical}.

\section{Implementation Details}
\label{app:implement}

\subsection{Training objective of Dr. MAMR}
The training objective for Dr. MAMR builds on Eq.~\ref{eq:rema} with two key modifications: 
(i) we remove the $\tfrac{1}{T_i}$ normalization over the number of turns, and 
(ii) we replace the step-level advantage with the weighted formulation in Eq.~\ref{eq:final}. 
The resulting objective is defined as follows:

\begin{equation}
\begin{aligned}
\label{eq:drmamr}
&\mathcal{J}_{\text{Dr.~MAMR}}(\theta) 
= \mathbb{E}_{(\mathbf{x},\mathbf{y}^*)\sim\mathcal{D},\, \{(\mathbf{m}_i, \mathbf{y}_i)\}_{i=1}^G \sim \pi_{\theta_{\mathrm{old}}}(\cdot\mid\mathbf{x})} 
\\
&\Bigg[
\frac{1}{G} \sum_{i=1}^G \sum_{t=1}^{T_i} 
\frac{1}{|\mathbf{y}_{i,t}|} \sum_{j=1}^{|\mathbf{y}_{i,t}|}
\Big(
\min\big(r_{i,t}(\theta) A^{\mathrm{step}}_{i,t},\, \mathrm{clip}(r_{i,t}(\theta), 1-\epsilon, 1+\epsilon) A^{\mathrm{step}}_{i,t}\big)
- \beta\, D_{\mathrm{KL}}(\pi_\theta \| \pi_{\mathrm{ref}})
\Big)
\Bigg],
\end{aligned}
\end{equation}
where the step-level advantage is given by Eq.~\ref{eq:final}:
\[
A^{\mathrm{step}}_{i,t} = \tilde{A}_{i,t} + \alpha \tilde{C}_{i,t} + \beta \tilde{R}_{i,t}.
\]

\subsection{Training on DeepscaleR}
We conduct all experiments using the Verl RL framework \citep{sheng2024hybridflow}. 
Given the substantial computational cost, we fix the hyperparameters at 
$\alpha = \beta = 0.1$ across all experiments, as this setting provides stable performance. 
We use bfloat16 precision for training, with a batch size of 128 and 128 sampled 
rollouts per training step.

\subsection{Shapley-inspired causal influence}
To group semantically similar steps for causal influence estimation, we use Qwen2.5-0.5B \footnote{\texttt{https://huggingface.co/Qwen/Qwen2.5-0.5B}} 
as the embedding model. Each step $s_{i,t}$ in a trajectory is encoded into a dense 
vector representation, and semantic similarity between steps is measured using 
cosine similarity. For each anchor step, we form a group $G_S(s_{i,t})$ by including 
all steps whose embeddings have cosine similarity of at least $0.9$ with the anchor. 
This threshold ensures that only highly similar steps---those expressing essentially 
the same idea, regardless of minor wording differences---are grouped together.

\subsection{Cold start for meta-think and reasoning}
For the experiments in Table~\ref{tab:main}, we adopt RL-from-base since the Qwen2.5B-Instruct family \citep{qwen2.5} demonstrates strong instruction-following capability. Moreover, as shown in ReMA~\citep{wan2025rema}, the performance gap between RL-from-base and RL-from-SFT is marginal.

\subsection{Cold start for restart behavior}
However, it is difficult for the base model to exhibit restart behavior directly. To address this, we collect expert data and apply supervised fine-tuning (SFT) to the base model. Specifically, building on the meta-thinking and reasoning datasets collected by ReMA \citep{wan2025rema, ye2025limo}, we use GPT-4o \citep{hurst2024gpt} to adversarially insert a few steps of noisy reasoning, followed by a restart, in order to simulate scenarios where the reasoning agent becomes lost in the conversation and the restart behavior enables recovery. For supervised fine-tuning, we use the LlamaFactory codebase, training the model for 3 epochs with a learning rate of 1e-5, a cosine learning rate scheduler, and a batch size of 8. We employ DeepSpeed ZeRO-2 for distributed training.

\section{More Experimental Details}
\label{app:expdetail}
% \subsection{More details of settings of preliminary experiments}
% \label{sec:pre_exp_set}

\subsection{Preliminary experiments on causal influence}
\label{sec:pre_causal}
\subsubsection{Experimental setup}
\label{sec:pre_exp_setting}
We follow the setting in \citep{wan2025rema} and train models on 7.5k training samples in MATH \citep{hendrycks2021measuring} and test on datasets: GSM8K \citep{cobbe2021training}, AIME24\footnote{\texttt{https://huggingface.co/datasets/AI-MO/aimo-validation-aime}}, AMC23\footnote{\texttt{https://huggingface.co/datasets/AI-MO/aimo-validation-amc}}, GaoKao2023En \citep{zhang2023evaluating}, Minerva Math \citep{lewkowycz2022solving}, and Olympiad Bench \citep{he2024olympiadbench}. We generate reasoning processes on the evaluation benchmark and subsequently measure the causal influence of each response following the descriptin in Sec. \ref{sec:preliminary}. 
% \zhiwei{how the causal influence calculated? not demonstrated in details (in our preliminary experiments, we measure the sequence i's causal influence on sequence i+1)}

\subsubsection{Prompts}
\begin{tcolorbox}[title=System Prompts for Meta-Think and Reasoning Agents of ReMA, colback=gray!5!white, colframe=black!75!white, fonttitle=\bfseries]
\small
\textbf{META-THINK AGENT SYSTEM PROMPT}

You are a meta-think agent that represents human high-level thinking processes. When solving a question, you will have a discussion with a human. Each time, think about what to do next. For example:
\begin{itemize}
    \item Exploring multiple angles and approaches
    \item Breaking down the solution into clear steps
    \item Continuously reflecting on intermediate results honestly and adapting your strategy as you progress
    \item Backtracking when necessary
    \item Requesting exploration of multiple solutions individually
    \item Finally, confirm the answer with the tag \texttt{[FINISH]}
\end{itemize}

\vspace{1em}
\textbf{REASONING AGENT SYSTEM PROMPT}

Please reason step by step following the given instruction. When asked to finalize your answer, put your answer within \verb|\boxed{}|.

\end{tcolorbox}

\begin{tcolorbox}[title=Refined System Prompts for Meta-Think and Reasoning Agents, colback=gray!5!white, colframe=black!75!white, fonttitle=\bfseries]
\small
\textbf{META-THINK AGENT SYSTEM PROMPT}

You are a meta-think agent that represents human high-level think process. When solving a question, you will have a discussion with the human, and each time you will think about what to do next. For example:
\begin{itemize}
    \item Exploring multiple angles and approaches
    \item Breaking down the solution into clear steps
    \item Continuously reflecting on intermediate results honestly and adapting your strategy as you progress
    \item Backtracking when necessary
    \item Requesting exploration of multiple solutions individually
    \item Finally, confirm the answer with the tag \texttt{[FINISH]}.
\end{itemize}

\vspace{0.5em}
\textbf{Please \emph{do not} focus on completing the task by calculating the final answer}; that step will be handled by a separate reasoning agent.

% \vspace{0.5em}

\vspace{1em}
\textbf{REASONING AGENT SYSTEM PROMPT}

You are a reasoning agent that follows structured problem-solving instructions step by step. 
% You are also an expert problem solver in a multi-turn conversation where information may be revealed gradually. 
Your goals are:

\begin{itemize}
    \item Follow the given instruction precisely.
    \item Reason step by step toward a solution.
    \item Avoid producing empty or blank outputs at any step.
    \item If uncertain, provide your best reasoning and partial answer rather than outputting nothing.
    \item Always provide a meaningful and non-empty response, even during intermediate steps.
    \item When you receive the signal \texttt{[FINISH]}, finalize your answer and place it within \verb|\boxed{}|.
    \item If unable to finalize, explain why and still output your best available answer within \verb|\boxed{}|.
\end{itemize}
\textbf{Remember: You must never produce trivial outputs.}

% \vspace{0.5em}
% \textbf{Deliberation Phase:}

% Before producing any solution, you must enter a deliberation phase and output:

% \begin{quote}
% \texttt{\textless pause\textgreater} \\
% \texttt{Decision: \textless Continue from previous answer | Aggregate instructions from all previous turns and restart\textgreater} \\
% \texttt{Reason: \textless one-sentence justification\textgreater}
% \end{quote}

% Your goal at each turn is to decide whether to \textbf{reuse your previous answer} or \textbf{restart by aggregating instructions from all previous turns}, in order to maximize correctness and consistency. You have an \textbf{unknown mode}: if you believe you cannot achieve the next goal perfectly using your previous answer, you must discard it and instead synthesize all relevant instructions and context accumulated so far before proceeding.

% \vspace{0.5em}
% \textbf{Rethinking Rules for New Information:}
% \begin{enumerate}[leftmargin=1.5em]
%     \item When new information arrives, re-examine all relevant previous assumptions and calculations.
%     \item If the new info contradicts or changes prior work, explicitly acknowledge this and recalculate from scratch.
%     \item Do not simply append new info to old reasoning — check if the prior approach is now invalid.
%     \item Ask specific clarifying questions only when essential facts are missing.
%     \item When revising due to new info, state: \textit{"Given this new information, I need to revise my approach."}
% \end{enumerate}

\end{tcolorbox}

\subsection{Preliminary experiments on how deliberation helps multi-turn reasoning}
\label{sec:pre_deliberation_appendix}
In this section, we provide additional details about the preliminary experiments referenced in Sec.~\ref{sec:deliberation}, which demonstrate that deliberation can enhance performance in multi-turn reasoning tasks.

% \subsubsection{Experimental Setting}
We adopt the experimental setup outlined in Sec.~\ref{sec:pre_exp_setting} to train both the meta-thinking agent and the reasoning agent using the original ReMA framework \citep{wan2025rema}. To evaluate the impact of deliberation, we introduce a variant of ReMA at inference time by modifying the system prompt of the reasoning agent as follows:
\begin{tcolorbox}[title=System Prompt for Reasoning Agent of ReMA+, colback=gray!5!white, colframe=black!75!white, fonttitle=\bfseries]
Please reason step by step. When asked to finalize your answer, place your result within \verb|\boxed{}|.

\medskip
\textbf{Deliberation Instruction}

If your previous reasoning is based on assumptions that are contradicted or made incomplete by new instructions, you must explicitly output:

\texttt{<restart>}\\
Reason: \texttt{<One-sentence justification>}  

\medskip
\textbf{Rethinking Rules for New Information}
\begin{enumerate}
    \item Upon receiving new information, re-evaluate all prior assumptions, steps, and conclusions.
    \item If the new information invalidates or alters your earlier reasoning, clearly acknowledge this and issue a \texttt{<restart>}.
    \item Do not simply add new information to prior reasoning — verify that the existing logic still holds.
    \item Ask clarifying questions only if essential information is missing and further reasoning is blocked.
    \item If you revise your reasoning due to new information, explicitly state: \\
    \texttt{"Given this new information, I need to revise my approach."}
\end{enumerate}
\end{tcolorbox}

% \section{Process Reward Fails to Mitigate the Lazy Agent Issue}
% In this section, we present experimental results showing that using a process reward model to assign credit at each turn fails to mitigate the lazy agent issue in ReMA.
% We follow the experimental setup described in Section~\ref{sec:pre_causal}, and adopt \texttt{Qwen/Qwen2.5-Math-PRM-7B}\footnote{\href{https://huggingface.co/Qwen/Qwen2.5-Math-PRM-7B}{Qwen/Qwen2.5-Math-PRM-7B}}
%  as the process reward model (PRM). We revise Eq.~\ref{eq:turn_grpo} as follows:
% \begin{equation}
% r_{i,t}(\theta) = r_{i,t}(\theta) + a_{i,t},
% \end{equation}
% where $a_{i,t}$ denotes the process reward for the $t$-th turn in the $i$-th rollout, provided by the PRM.

% The training curve of this approach is shown in Fig.~\ref{fig:prm_fail}. As observed, the model collapses rapidly after only 30 training steps. We attribute this to reward hacking, a well-known failure mode in RLHF settings, as documented in prior work \citep{zhang2025bradley, gao2023scaling, yang2024regularizing, li2025ruleadapter}.

% \begin{figure}
%     \centering
%     \includegraphics[width=0.65\linewidth]{figs/prm_fail.pdf}
%     \caption{Training curve of ReMA with process reward assigned for each turn.}
%     \label{fig:prm_fail}
% \end{figure}

\section{Case Study on Lazy Behavior}
\label{appednix:casestudy}

\begin{tcolorbox}[title=Case Study, colback=gray!5!white, colframe=gray!75!black]
\small
\texttt{[Question]}\\
Determine the number of ways to arrange the letters of the word 'ELLIPSE'. \textbf{[Groundtruth: 1260]}
\begin{tcolorbox}[title=ReMA, colback=gray!5!white, colframe=gray!90!black]
% \texttt{**ReMA**}\\
\textbf{Meta-thinking:} \\
Let's start by considering the total number of letters and any repetitions. The word ELLIPSE has 7 letters, with L repeated twice.
To find the number of distinct arrangements, we use the formula for permutations of a multiset:
\[
\frac{n!}{n_1! \cdot n_2! \cdot \ldots \cdot n_k!}
\]

For ELLIPSE:
- \( n = 7 \)
- L appears 2 times
- E, I, P, S each appear once
\[
\frac{7!}{2!} = \frac{5040}{2} = 2520
\]
\textbf{Reasoning:}
\[
\boxed{ }
\]
\textbf{Meta-thinking:}\\
\[\text{[FINISH]}\] 
\textbf{Reasoning:}
\[
\boxed{2520}
\]
\end{tcolorbox}
% \begin{tcolorbox}[title=Dr. MAMR, colback=gray!5!white, colframe=gray!90!black]
% % \texttt{**Ours**}\\
% \textbf{Meta-thinking:} \\
% To determine the number of ways to arrange the letters of the word "ELLIPSE", we need to consider the total number of letters and any repetitions of letters.

% 1. First, let's count the total number of letters and identify any repeated letters.\\
% 2. Then, we'll use the formula for permutations of a multiset.

% Let's start by counting the letters. How many letters are there in the word "ELLIPSE"? And which letters are repeated?\\
% \textbf{Reasoning:} \\
% The word "ELLIPSE" consists of 7 letters: E, L, L, I, P, S, E.
% We can see that the letter 'E' is repeated twice, and the letter 'L' is also repeated twice. The other letters, I, P, and S, each appear once.
% To determine the number of distinct arrangements of these letters, we use the formula for permutations of a multiset:
% \[
% \frac{n!}{n_1! \cdot n_2! \cdot \ldots \cdot n_k!}
% \]
% For the word "ELLIPSE":
% - \( n = 7 \)
% - 'E' occurs 2 times, 'L' occurs 2 times, others occur once
% \[
% \frac{7!}{2! \cdot 2! \cdot 1! \cdot 1! \cdot 1!} = \frac{5040}{4} = 1260
% \]
% \[
% \boxed{1260}
% \]
% \textbf{Meta-thinking:} \\
% Great job walking through the calculation! Let's summarize:
% \begin{itemize}
%   \item 7 letters in total
%   \item 'E' and 'L' both appear twice
%   \item Used multiset formula correctly
%   \item Answer: \( \boxed{1260} \)
% \end{itemize}
% % \[\text{[FINISH]}\]
% % \textbf{Reasoning:} \\

% \end{tcolorbox}
\end{tcolorbox}

\begin{tcolorbox}[title=Dr. MAMR, colback=gray!5!white, colframe=gray!90!black]
% \texttt{**Ours**}\\
\textbf{Meta-thinking:} \\
To determine the number of ways to arrange the letters of the word "ELLIPSE", we need to consider the total number of letters and any repetitions of letters.

1. First, let's count the total number of letters and identify any repeated letters.\\
2. Then, we'll use the formula for permutations of a multiset.

Let's start by counting the letters. How many letters are there in the word "ELLIPSE"? And which letters are repeated?\\
\textbf{Reasoning:} \\
The word "ELLIPSE" consists of 7 letters: E, L, L, I, P, S, E.
We can see that the letter 'E' is repeated twice, and the letter 'L' is also repeated twice. The other letters, I, P, and S, each appear once.
To determine the number of distinct arrangements of these letters, we use the formula for permutations of a multiset:
\[
\frac{n!}{n_1! \cdot n_2! \cdot \ldots \cdot n_k!}
\]
For the word "ELLIPSE":
- \( n = 7 \)
- 'E' occurs 2 times, 'L' occurs 2 times, others occur once
\[
\frac{7!}{2! \cdot 2! \cdot 1! \cdot 1! \cdot 1!} = \frac{5040}{4} = 1260
\]
\[
\boxed{1260}
\]
\textbf{Meta-thinking:} \\
Great job walking through the calculation! Let's summarize:
\begin{itemize}
  \item 7 letters in total
  \item 'E' and 'L' both appear twice
  \item Used multiset formula correctly
  \item Answer: \( \boxed{1260} \)
\end{itemize}
% \[\text{[FINISH]}\]
% \textbf{Reasoning:} \\

\end{tcolorbox}

\section{Proofs}

Following Eq.\ref{eq:rema}, ReMA's training objective (per batch) is:
\begin{equation}
\begin{aligned}
\mathcal{J}(\theta) = \mathbb{E}\Bigg[
\frac{1}{G} \sum_{i=1}^G \frac{1}{T_i} \sum_{t=1}^{T_i} \frac{1}{|y_{i,t}|} \sum_{j=1}^{|y_{i,t}|} \left(
\min\{u_{i,t,j}, v_{i,t,j}\} - \beta\, D_{\mathrm{KL}}(\pi_\theta \,\|\, \pi_{\mathrm{ref}})
\right)
\Bigg],
\end{aligned}
\end{equation}
where
\begin{align}
u_{i,t,j} &= r_{i,t}(\theta)\, \hat{A}_{i,t,j}, \\
v_{i,t,j} &= \mathrm{clip}(r_{i,t}(\theta), 1{-}\epsilon, 1{+}\epsilon)\, \hat{A}_{i,t,j}.
\end{align}

Turn-level importance ratio:
\begin{equation}
r_{i,t}(\theta) = \frac{1}{|y_{i,t}|} \sum_{j'=1}^{|y_{i,t}|} 
\underbrace{
\frac{\pi_\theta(y_{i,t,j'} \mid c_{i,t,j'})}{\pi_{\theta_{\mathrm{old}}}(y_{i,t,j'} \mid c_{i,t,j'})}
}_{=:~r_{i,t,j'}(\theta)},
\end{equation}
where the context is:
\[
c_{i,t,j'} := (x_i, \{m_{i,\cdot}, y_{i,\cdot}\}_{<t}, m_{i,t}, y_{i,t,<j'}).
\]

We derive the turn-$t$ contribution for a fixed trajectory $i$, i.e.,
\begin{equation}
L_{i,t}(\theta)
:= \frac{1}{T_i} \cdot \frac{1}{|y_{i,t}|} \sum_{j=1}^{|y_{i,t}|}
\min\{u_{i,t,j}, v_{i,t,j}\}
- \frac{\beta}{T_i} \cdot \frac{1}{|y_{i,t}|} \sum_{j=1}^{|y_{i,t}|}
D_{\mathrm{KL}}\left(\pi_\theta(\cdot \mid c_{i,t,j}) \,\|\, \pi_{\mathrm{ref}}(\cdot \mid c_{i,t,j})\right).
\end{equation}

% \paragraph{2) Exact gradient without clipping and without KL}

Without considering the KL divergence and clipping,
\begin{equation}
L_{i,t}(\theta) = \frac{1}{T_i} \cdot \frac{1}{|y_{i,t}|} \sum_j r_{i,t}(\theta)\, \hat{A}_{i,t,j}
= \frac{1}{T_i} \cdot \bar{A}_{i,t} \cdot r_{i,t}(\theta),
\end{equation}
where
\begin{equation}
\bar{A}_{i,t} := \frac{1}{|y_{i,t}|} \sum_{j=1}^{|y_{i,t}|} \hat{A}_{i,t,j}.
\end{equation}

Compute the gradient:
% \begin{equation}
% \nabla_\theta r_{i,t}(\theta)
% = \frac{1}{|y_{i,t}|} \sum_{j'=1}^{|y_{i,t}|}
% \nabla_\theta r_{i,t,j'}(\theta)
% = \frac{1}{|y_{i,t}|} \sum_{j'=1}^{|y_{i,t}|}
% r_{i,t,j'}(\theta) \cdot \nabla_\theta \log \pi_\theta(y_{i,t,j'} \mid c_{i,t,j'}).
% \end{equation}

\begin{equation}
\begin{aligned}
\nabla_\theta r_{i,t}(\theta)
&= \nabla_\theta \left( \frac{1}{|y_{i,t}|} \sum_{j'=1}^{|y_{i,t}|} r_{i,t,j'}(\theta) \right) \\[6pt]
&= \frac{1}{|y_{i,t}|} \sum_{j'=1}^{|y_{i,t}|} \nabla_\theta 
\left( \frac{\pi_\theta(y_{i,t,j'} \mid c_{i,t,j'})}
{\pi_{\theta_{\mathrm{old}}}(y_{i,t,j'} \mid c_{i,t,j'})} \right) \\[6pt]
&= \frac{1}{|y_{i,t}|} \sum_{j'=1}^{|y_{i,t}|} 
\frac{1}{\pi_{\theta_{\mathrm{old}}}(y_{i,t,j'} \mid c_{i,t,j'})}
\nabla_\theta \pi_\theta(y_{i,t,j'} \mid c_{i,t,j'}) \\[6pt]
&= \frac{1}{|y_{i,t}|} \sum_{j'=1}^{|y_{i,t}|} 
\frac{\pi_\theta(y_{i,t,j'} \mid c_{i,t,j'})}
{\pi_{\theta_{\mathrm{old}}}(y_{i,t,j'} \mid c_{i,t,j'})}
\nabla_\theta \log \pi_\theta(y_{i,t,j'} \mid c_{i,t,j'}) \\[6pt]
&= \frac{1}{|y_{i,t}|} \sum_{j'=1}^{|y_{i,t}|}
r_{i,t,j'}(\theta)\, \nabla_\theta \log \pi_\theta(y_{i,t,j'} \mid c_{i,t,j'}).
\end{aligned}
\end{equation}

Therefore, the exact per-turn gradient (no clipping, no KL) is:
\begin{equation}
% \boxed{
\nabla_\theta L_{i,t}(\theta)
= \frac{1}{T_i} \cdot \bar{A}_{i,t} \cdot
\left( \frac{1}{|y_{i,t}|} \sum_{j'=1}^{|y_{i,t}|}
r_{i,t,j'}(\theta)\, \nabla_\theta \log \pi_\theta(y_{i,t,j'} \mid c_{i,t,j'})
\right)
% }
\end{equation}

% This shows the turn-level coupling: the mean advantage $\bar{A}_{i,t}$ multiplies a ratio-weighted mean of token log-prob gradients.

% If you prefer a “per-token” view, note there is no separate sum over $j$ and $j'$: the ratio’s gradient already averages over all tokens in the turn (indexed by $j'$), multiplied by the turn-mean advantage $\bar{A}_{i,t}$.
We define the aggregated turn-$t$ stochastic contribution as
\begin{equation}
Z_t(\tau) \;\triangleq\; \frac{1}{|y_t|} \sum_{j=1}^{|y_t|}
r_t(\theta)\,\hat{A}_{t,j}\,\nabla_\theta \log \pi_\theta\!\left(y_{t,j}\mid x,m_{\le t},y_{<t},y_{t,<j}\right).
\end{equation}
Then the ReMA gradient contribution at turn $t$ is
\begin{equation}
g_t(\tau) \;=\; \frac{1}{T(\tau)}\,Z_t(\tau).
\end{equation}

Let
\[
\kappa \;\triangleq\; 
\frac{\|Z_t(\tau^{\mathrm L})\|}
     {\|Z_t(\tau^{\mathrm S})\|}.
\]
Then the relative gradient magnitude satisfies
\[
\frac{\|g_t(\tau^{\mathrm S})\|}
     {\|g_t(\tau^{\mathrm L})\|}
\;=\; \frac{T_{\mathrm L}}{T_{\mathrm S}} \cdot \frac{1}{\kappa}.
\]
In particular, if $\kappa < \tfrac{T_{\mathrm L}}{T_{\mathrm S}}$, then 
\[
\frac{\|g_t(\tau^{\mathrm S})\|}
     {\|g_t(\tau^{\mathrm L})\|} > 1.
\]
% so the shorter trajectory $\tau^{\mathrm S}$ receives a strictly larger gradient magnitude than the longer trajectory $\tau^{\mathrm L}$.
% \end{proposition}

\section{Preliminary Experiments for Theoretical Analysis}
\label{app:expforturn}
\begin{figure}[h]
    \centering
    \includegraphics[width=0.5\linewidth]{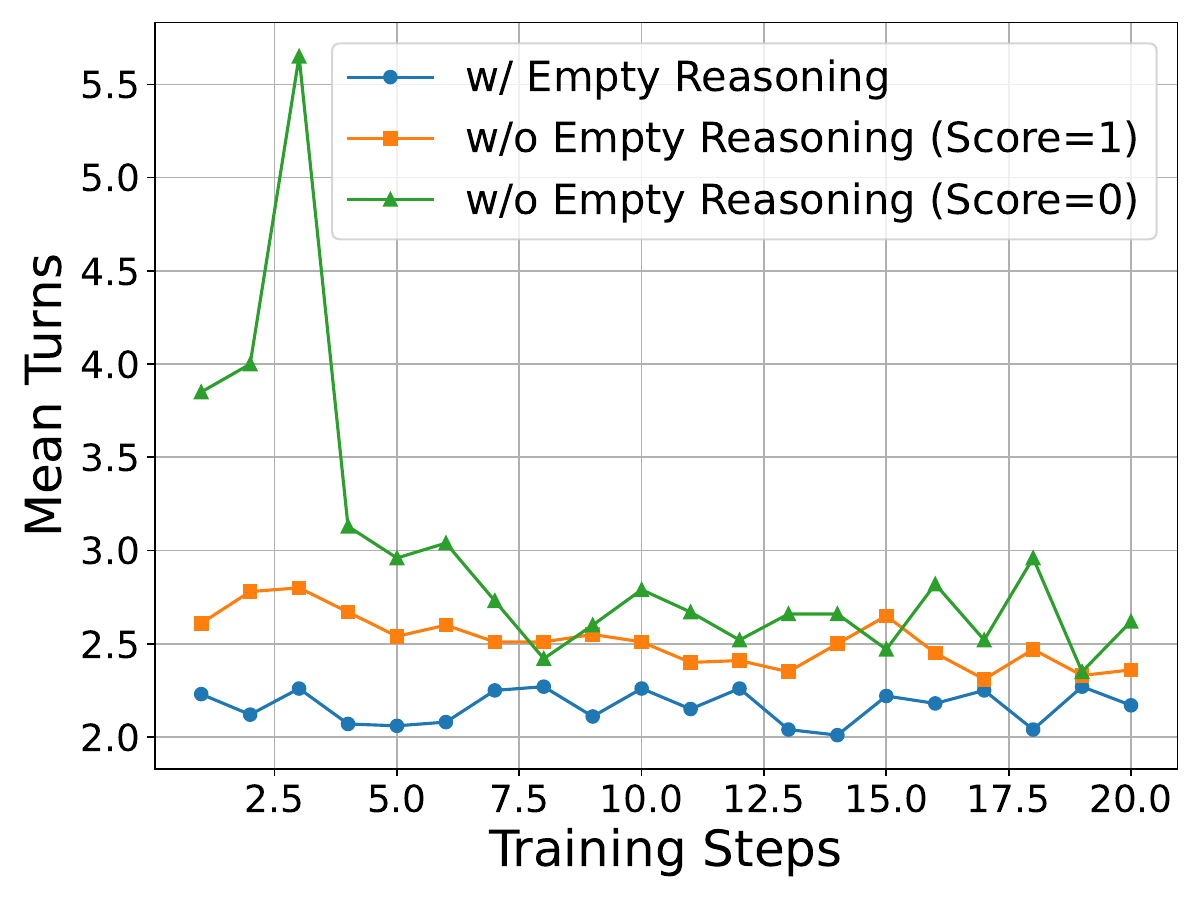}
    \caption{Mean number of turns comparing reasoning processes w/ and w/o lazy-agent behavior.}
    \label{fig:premeanturn}
\end{figure}

We present preliminary results on the mean number of turns in the reasoning process for cases with empty outputs and trivially copy the other's response (i.e., reasoning process exhibiting lazy-agent behavior) and those without empty outputs (i.e., reasoning process without lazy agents). We report results from the first 20 training steps, as this initial stage is critical in shaping the agent’s behavior. As shown in Fig.~\ref{fig:premeanturn}, the number of turns for reasoning process exhibiting lazy-agent behavior is consistently smaller than that of non-lazy agents.

\section{Process Reward Fails to Mitigate the Lazy Agent Issue}
In this section, we present experimental results showing that using a process reward model to assign credit at each turn fails to mitigate the lazy agent issue in ReMA.
We follow the experimental setup described in Section~\ref{sec:pre_causal}, and adopt \texttt{Qwen/Qwen2.5-Math-PRM-7B}\footnote{\href{https://huggingface.co/Qwen/Qwen2.5-Math-PRM-7B}{Qwen/Qwen2.5-Math-PRM-7B}}
 as the process reward model (PRM). We revise Eq.~\ref{eq:turn_grpo} as follows:
\begin{equation}
r_{i,t}(\theta) = r_{i,t}(\theta) + a_{i,t},
\end{equation}
where $a_{i,t}$ denotes the process reward for the $t$-th turn in the $i$-th rollout, provided by the PRM.

\begin{figure}
    \centering
    \includegraphics[width=0.65\linewidth]{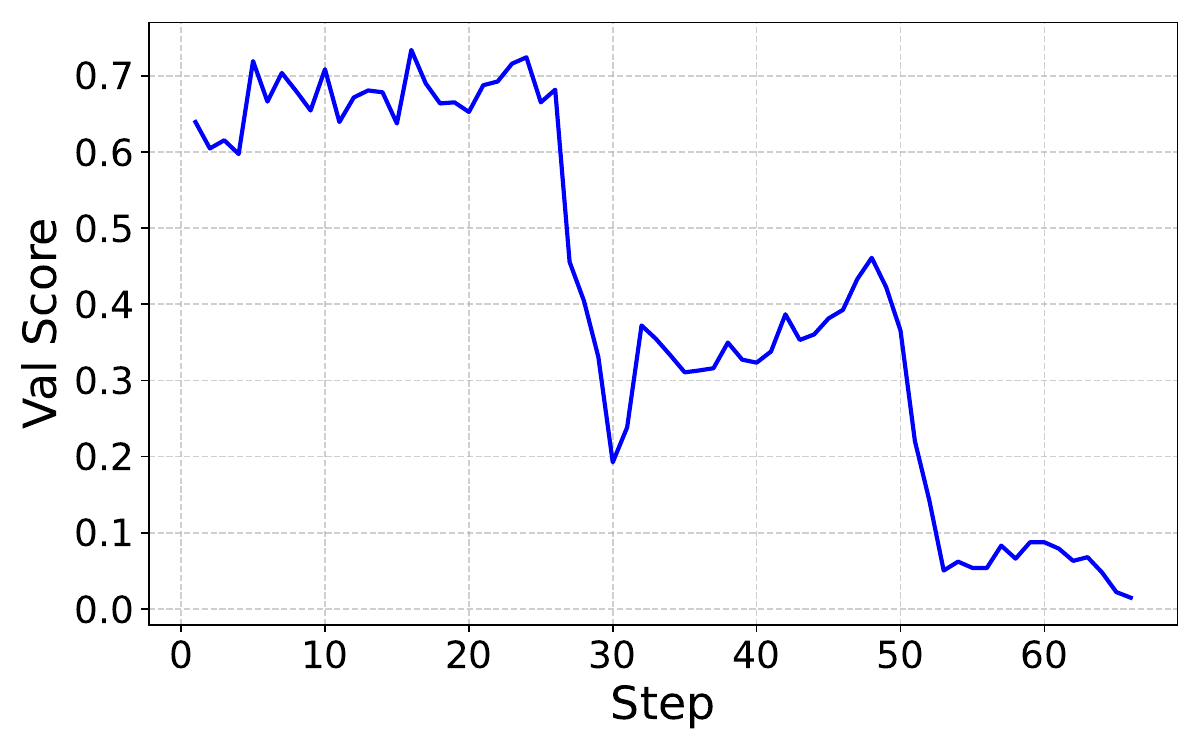}
    \caption{Training curve of ReMA with process reward assigned for each turn.}
    \label{fig:prm_fail}
\end{figure}

The training curve of this approach is shown in Fig.~\ref{fig:prm_fail}. As observed, the model collapses rapidly after only 30 training steps. We attribute this to reward hacking, a well-known failure mode in RLHF settings, as documented in prior work \citep{zhang2025bradley, gao2023scaling, yang2024regularizing, li2025ruleadapter}.

\section{Additional Experiment Results}

\subsection{Results across benchmarks on 3B model}
\label{app:mainfor3b}

In this section, we present additional pass@1 performance results on the 3B model, as shown in Table~\ref{app:main3b}. From the table, we observe that our Dr. MAMR consistently outperforms both single-agent GRPO and ReMA. However, the performance gains are less pronounced on the 7B and 14B models. We attribute this to their weaker instruction-following capability, which limits the performance upper bound of the multi-agent system.

\begin{table}[h]
\centering
\caption{Performance on math benchmarks with 3B base model.}
\label{app:main3b}
\begin{tabular}{c|c|ccccc}
\toprule
\textbf{Model} & \textbf{Benchmark} & \textbf{GRPO} & \textbf{VRP (CoT)} & \textbf{ReMA} & \textbf{Dr. MAMR (Ours)} \\
\midrule
\multirow{8}{*}{\shortstack[l]{Qwen2.5\\-3B\\-Instruct}}
&MATH500          & 65.60 & 65.20  & 62.60 & 66.20\\
&GSM8K            & 85.30 & 72.02  & 83.17 & 85.37\\
&AIME24           & 13.33 & 3.33   & 3.33  & 16.67\\
&AMC23            & 40.00 & 20.00  & 42.50 & 50.00\\
&Gaokao2023en     & 54.30 & 30.91  & 52.73 & 55.33\\
&Minerva Math     & 31.20 & 16.91  & 26.47 & 32.35\\
&Olympiad Bench   & 30.20 & 6.07   & 27.56 & 30.57\\
\cmidrule(lr){2-6}
&\textbf{Average} &    45.70   &   30.63     &  42.62     & 48.07\\
\bottomrule
\end{tabular}
\end{table}

\subsection{Training curve comparison on the 3B Model}
\label{app:3Bcurve}

\begin{figure}[h]
    \centering
    \includegraphics[width=0.5\linewidth]{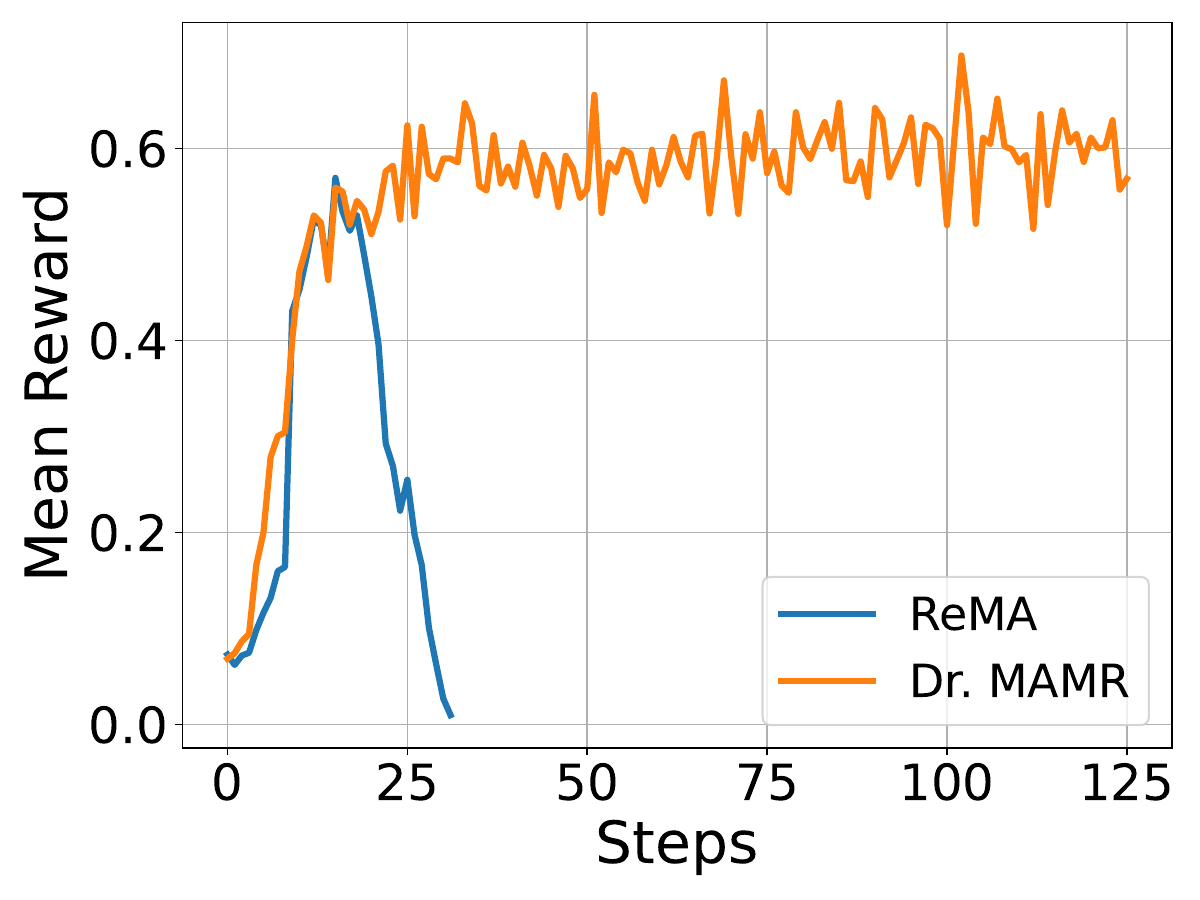}
    \caption{Training curves of 3B model trained with ReMA and Dr. MAMR.}
    \label{fig:3b_curve}
\end{figure}

To further examine the training stability of smaller base models, we provide additional experiments on the 3B model. In this setting, we use MATH as the training dataset to rule out the possibility that collapse is solely caused by overly difficult data. The training curves reporting mean reward are shown in Fig.~\ref{fig:3b_curve}.  

From the figure, we observe that under the ReMA framework, the 3B model collapses rapidly after only 20 training steps. In contrast, our Dr.~MAMR framework maintains stable progress throughout training. This highlights how critical the lazy-agent issue becomes when the base model is relatively weak. Importantly, even with a less capable model, Dr.~MAMR is still able to ensure stable training, underscoring its robustness.

\subsection{Case study on restart behavior}
\label{app:caserestart}
In this section, we present a case study on restart behavior, illustrating how restarts help the reasoning agent recover from its earlier mistakes. The full reasoning process is shown below. This case study demonstrates how restart behavior enables the agent to abandon
an incomplete reasoning trajectory and reinitiate a more systematic approach.
By explicitly reconsidering the intersection points, the agent successfully
recovered the correct quadrilateral and computed the ground-truth area of
$4\sqrt{3}$. Without such a restart, the agent would likely have produced an
incorrect or incomplete solution.

\begin{tcolorbox}[title=Case study on restart behavior, colback=gray!5!white, colframe=gray!75!black]

\texttt{[Question]}
Find the area of the quadrilateral formed by the intersection points of the parabolas $y^2 = 3x$, $y^2 = -3x$, and the circle $x^2 + y^2 = 4$. \textbf{[Groundtruth: $4\sqrt{3}$]}

% \begin{tcolorbox}[title=ReMA, colback=gray!5!white, colframe=gray!90!black]

\textbf{Meta-thinking:}\\
To find the area of the quadrilateral formed by the points of intersection between the parabolas and the circle, begin by determining the intersection points. Substitute $y^2=3x$ and $y^2=-3x$ into $x^2+y^2=4$ and work through the feasible solutions.

\textbf{Reasoning:}\\
For $y^2=3x$: $x^2+3x=4 \;\Rightarrow\; (x+4)(x-1)=0$, so $x=1$ or $-4$. But $x=-4$ gives $y^2=-12$. Thus $(1,\pm\sqrt{3})$.\\
For $y^2=-3x$: $x^2-3x=4 \;\Rightarrow\; (x-4)(x+1)=0$, so $x=4$ or $-1$. But $x=4$ gives $y^2=-12$. Thus $(-1,\pm\sqrt{3})$.

\textbf{Meta-thinking:}\\
The feasible intersections suggest four symmetric points on the circle: $(1,\pm\sqrt{3})$ and $(-1,\pm\sqrt{3})$. It’s natural to check the configuration these points form and then evaluate the area accordingly.

\textbf{Reasoning:}\\
\texttt{<restart>} Decision: Rethink from scratch. The feasible intersections are:  
- From $y^2 = 3x$: $(1,\pm\sqrt{3})$  
- From $y^2 = -3x$: $(-1,\pm\sqrt{3})$  

Thus, four points of intersection:  
$A=(1,\sqrt{3}), B=(1,-\sqrt{3}), C=(-1,\sqrt{3}), D=(-1,-\sqrt{3})$.

\textbf{Meta-thinking:} \\
These points form a rectangle.  
- Width: $|1-(-1)| = 2$  
- Height: $|\sqrt{3}-(-\sqrt{3})| = 2\sqrt{3}$  

Area:
\[
\text{Area} = 2 \times 2\sqrt{3} = 4\sqrt{3}
\]

[FINISH]

\textbf{Reasoning:} \\
\[
\boxed{4\sqrt{3}}
\]

% \end{tcolorbox}
\end{tcolorbox}

\subsection{Scaling}
\label{app:scaling}
In this section, we present the full results of test-time scaling on pass@K performance when comparing Dr. MAMR with ReMA, as shown in Fig.~\ref{app:scaling}. We observe that on harder benchmarks such as AIME24 and OlympiadBench, the performance gap between Dr. MAMR and ReMA widens as $K$ increases, while on relatively easier benchmarks, Dr. MAMR consistently achieves better performance. These results highlight the strong capability of Dr. MAMR in handling a wide range of reasoning tasks.

\begin{figure}
    \centering
    \includegraphics[width=\linewidth]{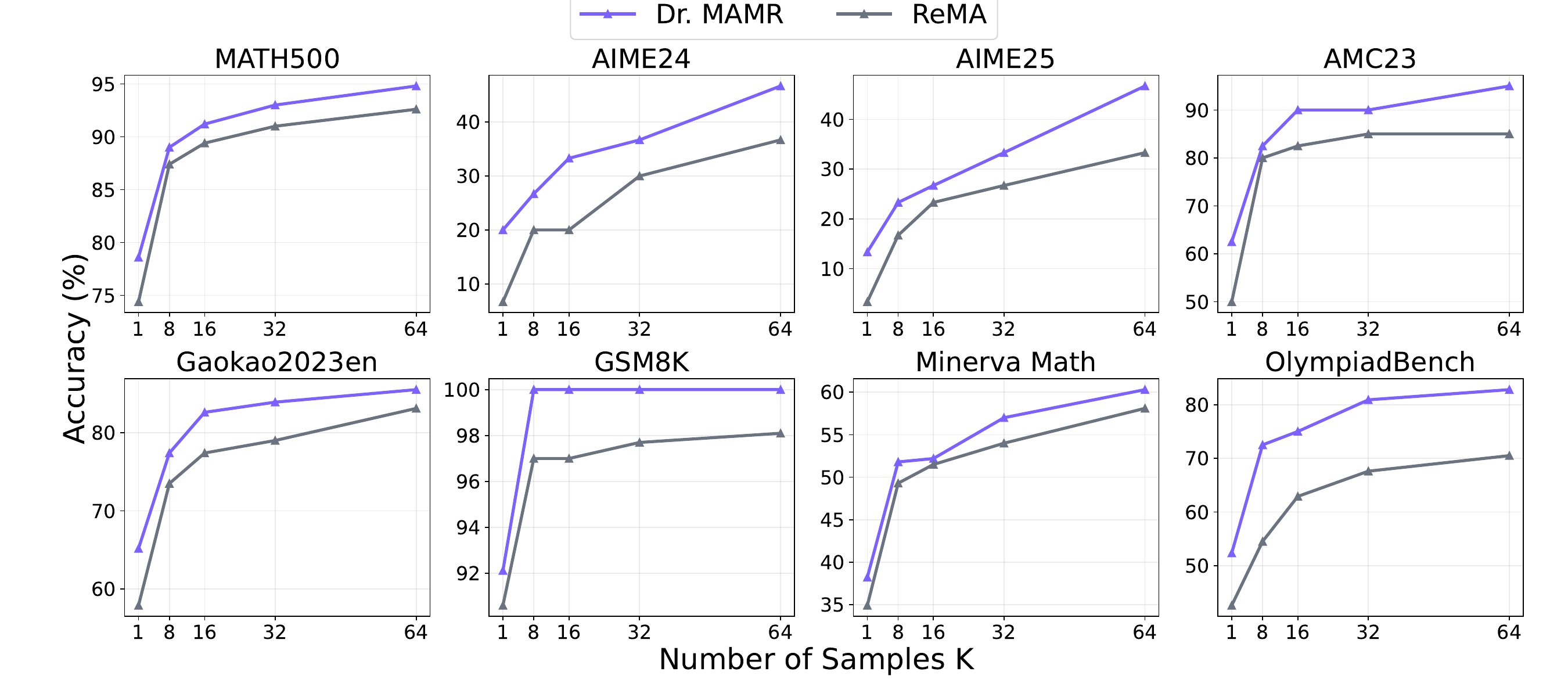}
    \caption{Pass@K performance.}
    \label{app:fullscaling}
\end{figure}

\section{The Use of Large Language Models}
Large Language Models (LLMs) were used to assist in writing and polishing this manuscript. Specifically, an LLM was employed to improve clarity, refine language, check grammar, and enhance overall readability.

The LLM was not involved in ideation, research design, data analysis, or the development of scientific content. All research concepts, methods, and analyses were independently conceived and conducted by the authors.

The authors take full responsibility for the manuscript’s content, including any text refined with LLM assistance. All LLM-generated content adheres to ethical standards and does not constitute plagiarism or scientific misconduct.

\end{document}